\documentclass{article}

\PassOptionsToPackage{numbers, compress}{natbib}
\usepackage[preprint]{neurips_2021}

\usepackage[utf8]{inputenc} % allow utf-8 input
\usepackage[T1]{fontenc}    % use 8-bit T1 fonts
\usepackage{hyperref}       % hyperlinks
\usepackage{url}            % simple URL typesetting
\usepackage{booktabs}       % professional-quality tables
\usepackage{amsfonts}       % blackboard math symbols
\usepackage{nicefrac}       % compact symbols for 1/2, etc.
\usepackage{microtype}      % microtypography
\usepackage{xcolor}         % colors
\usepackage[pdftex]{graphicx}
\usepackage{amsmath}
\usepackage{float}

\providecommand{\R}{\mathbb{R}}

\providecommand{\bb}{\mathbf{b}}

\providecommand{\ww}{\mathbf{w}}
\providecommand{\xx}{\mathbf{x}}
\providecommand{\yy}{\mathbf{y}}

\providecommand{\shared}{{\text{\tiny\textsf{shared}}}}

\providecommand{\inp}{{\text{\tiny\textsf{in}}}}
\providecommand{\out}{{\text{\tiny\textsf{out}}}}

\title{iFedAvg -- Interpretable Data-Interoperability\\ for Federated Learning}

\author{
  David Roschewitz \\
  ETH Zürich\\
  Switzerland \\
  \texttt{david.roschewitz@inf.ethz.ch} \\
  \And
   Mary-Anne Hartley \\
   EPFL, Lausanne \\
   Switzerland \\
   \texttt{mary-anne.hartley@epfl.ch} \\
   \AND
   Luca Corinzia \\
   ETH Zürich \\
   Switzerland \\
   \texttt{luca.corinzia@inf.ethz.ch} \\
   \And
   Martin Jaggi \\
   EPFL, Lausanne \\
   Switzerland \\
   \texttt{martin.jaggi@epfl.ch} \\
}

\begin{document}

\maketitle

\begin{abstract}
Recently, the ever-growing demand for privacy-oriented machine learning has motivated researchers to develop federated and decentralized learning techniques, allowing individual clients to train models collaboratively without disclosing their private datasets. However, widespread adoption has been limited in domains relying on high levels of user trust, where assessment of data compatibility is essential. In this work, we define and address low interoperability induced by underlying client data inconsistencies in federated learning for tabular data. The proposed method, \texttt{iFedAvg}, builds on federated averaging adding local element-wise affine layers to allow for a personalized and granular understanding of the collaborative learning process. Thus, enabling the detection of outlier datasets in the federation and also learning the compensation for local data distribution shifts without sharing any original data. We evaluate \texttt{iFedAvg} using several public benchmarks and a previously unstudied collection of real-world datasets from the 2014 - 2016 West African Ebola epidemic, jointly forming the largest such dataset in the world. In all evaluations, \texttt{iFedAvg} achieves competitive average performance with negligible overhead. It additionally shows substantial improvement on outlier clients, highlighting increased robustness to individual dataset shifts. Most importantly, our method provides valuable client-specific insights at a fine-grained level to guide interoperable federated learning.
\end{abstract}

\section{Introduction}
Institutions with sensitive data, such as hospitals, cannot typically share patient data due to privacy regulations. However, solely relying on in-house data can lead to models with poor generalization due to the limited, and potentially biased, input data. Federated learning (FL) partially addresses this issue by enabling various clients to contribute and benefit from a collaborative learning process without revealing their underlying data. In practice though, individual clients can benefit from the collaborative training only if their data is compatible with that of other participating institutions. This can lead to situations where the client data is not interoperable, and where joining the federated learning process yields no benefits or even has a detrimental effect. Furthermore, a lack of transparency of the federated learning process impairs the trust of the federation, limiting its adoption by more institutions.

Thus, there is a clear need to address the interoperability of current federated learning approaches. Such methods should not only detect potential data shifts and automatically correct them but, more importantly, they should also be easily interpretable for stakeholders to visually assess the suitability of the collaboration.

We study these challenges and propose a novel method for a setting inspired by a real-world medical dataset collected during the 2014-16 Ebola epidemic. The data was collected at different treatment centres, by various organizations in multiple countries generating an inherently heterogeneous dataset. In practice, no single agent would have access to the data of other agents, necessitating a federated learning approach. Further details are outlined in Section \ref{sec:ebola}. Our method is tailored to handle tabular datasets which are abundant in practice and for which feature-shifts are intuitive to understand.

The fundamental approach that we consider is to learn a personalized data transformation for each client during the federated training procedure. This transformation can be viewed as a local re-normalization or embedding that makes clients more interoperable without ever exchanging data. Through deliberate design, our method attains unparalleled transparency, allowing fine-grained interpretation of the learned shifts for each feature of each client relative to their peers in the federation. This ability to compare data shifts across clients means that interoperability can be easily assessed. 
For example, clients collecting data in pediatric, adult or geriatric medicine would not only have the "age" feature highlighted as responsible for their "outlier" status, but the directionality and relative magnitude of their outlier shifts can also be assessed.

Our main contributions are the following:
\begin{enumerate}
    \item We propose a novel framework, \texttt{iFedAvg} which detects and corrects interoperability issues in federated learning on tabular datasets.
    \item We present visualization tools for practitioners to assess the feature-wise compatibility to collaborative learning.
    \item Finally, we demonstrate the potential of the proposed method on a previously unstudied collection of data from the West African Ebola epidemic and multiple public benchmarks.
\end{enumerate}

\textbf{Outlook} \quad In Section \ref{sec:iFedAvg} we outline our method, and we present a detailed introduction to the Ebola dataset in Section \ref{sec:ebola}. Subsequently we present the experimental setup and results in Sections \ref{sec:experimentalsetup} and \ref{sec:results}. We provide a discussion and conclusions in the last section.  

\textbf{Related Work} \quad The concept of federated learning was formulated by \citet{mcmahan2017fl} as ``\emph{collaborative machine learning without centralized training data}'' alongside \textit{Federated Averaging} (\texttt{FedAvg}). The method proved efficient in learning a single, global, gradient-based model from many clients' data in a private fashion. However, this approach is impaired in the realistic setting with statistically heterogeneous client data \citet{zhao2018federated}. In order to address some of these shortcomings, multiple extensions have been proposed. For instance, \texttt{FedProx} by \citet{li2018fedprox} allows for inexact local computations and regularizes client drift using an additive proximal term in the loss function. \citet{karimireddy2020scaffold} introduced \texttt{SCAFFOLD}, which explicitly uses control variates to improve convergence. 
From an optimizer perspective, SGD with server-side momentum has proven effective~\cite{hsu2019fedavgM}, a result further investigated by \citet{reddi2020AFO} that proposes the use of adaptive optimizers for federated learning such as \texttt{FedAdam}. For a more comprehensive overview, we refer to the excellent review of federated learning by \citet{kairouz2019advances}.

Learning a client-specific \textit{data transformation}, as we propose here, can be seen as a special case of training personalized models for each client as opposed to a single global model. In the area of personalized federated learning (PFL), various approaches have been identified. A relevant selection includes transfer learning, multi-task learning, meta-learning and personalization layers. Transfer and meta-learning focus on tuning an initial model, usually the global federated model, to each client. Techniques here include fine-tuning \citet{yu2020FT} and Model-Agnostic-Meta-Learning (MAML) \citet{jiang2019improving}. \citet{fallah2020personalizedmetahessian} devise \texttt{Per-FedAvg}, leveraging a second-order derivative to account for personalization throughout the process. Multi-task learning approaches focus on jointly learning multiple models for a variety of tasks with different levels of similarity. This can be applied to PFL, as in MOCHA \citep{smith2017federated}, or using a Bayesian framework \citep{corinzia2019variational}. Furthermore, personalization can be achieved by client-specific layers, as proposed by~\cite{arivazhagan2019personalizationlayers}, who show that locally trained output layers are effective for image classification. Furthermore, \citet{deng2020APFL} propose a method, \texttt{APFL}, which actively optimizes a global and a local model, and blends the two models concurrently. The summary overview provided by \citet{kulkarni2020surveypersonalization} includes additional approaches for PFL.

Understanding \textit{differences} between clients and interpreting the federated learning process as a whole is, to the best of our knowledge, a problem not previously investigated. Standard neural-network model-interpretation techniques such as LIME \cite{ribeiro2016LIME} are, unfortunately, not immediately applicable to federated learning and thus would not fulfil our objectives. For vertically (i.e., feature-wise) partitioned data and tree-based models, SHAP-values have been investigated by \citet{zheng2020vertical} and \citet{wang2019vertshap}, but the approach is not applicable to our setting as we focus on a \textit{horizontally} (i.e., sample-wise) partitioned dataset. Furthermore, SHAP values learn feature-wise contributions to the model, rather than learning and compensating for potential data biases.  \citet{imakura2021interpretablecollabanalysis} present an "interpretable non-model sharing collaborative data analysis method as a federated learning system". While retaining privacy, the authors assume having access to a shared public \textit{anchor} dataset and focus on model-interpretability only. Hence the need for a method allowing for interpretability at the process level enabling comparisons between clients.

\section{Interpretable and data-interoperable federated averaging}
\label{sec:iFedAvg}

\begin{figure}
  \centering
  \includegraphics[width=0.99\linewidth]{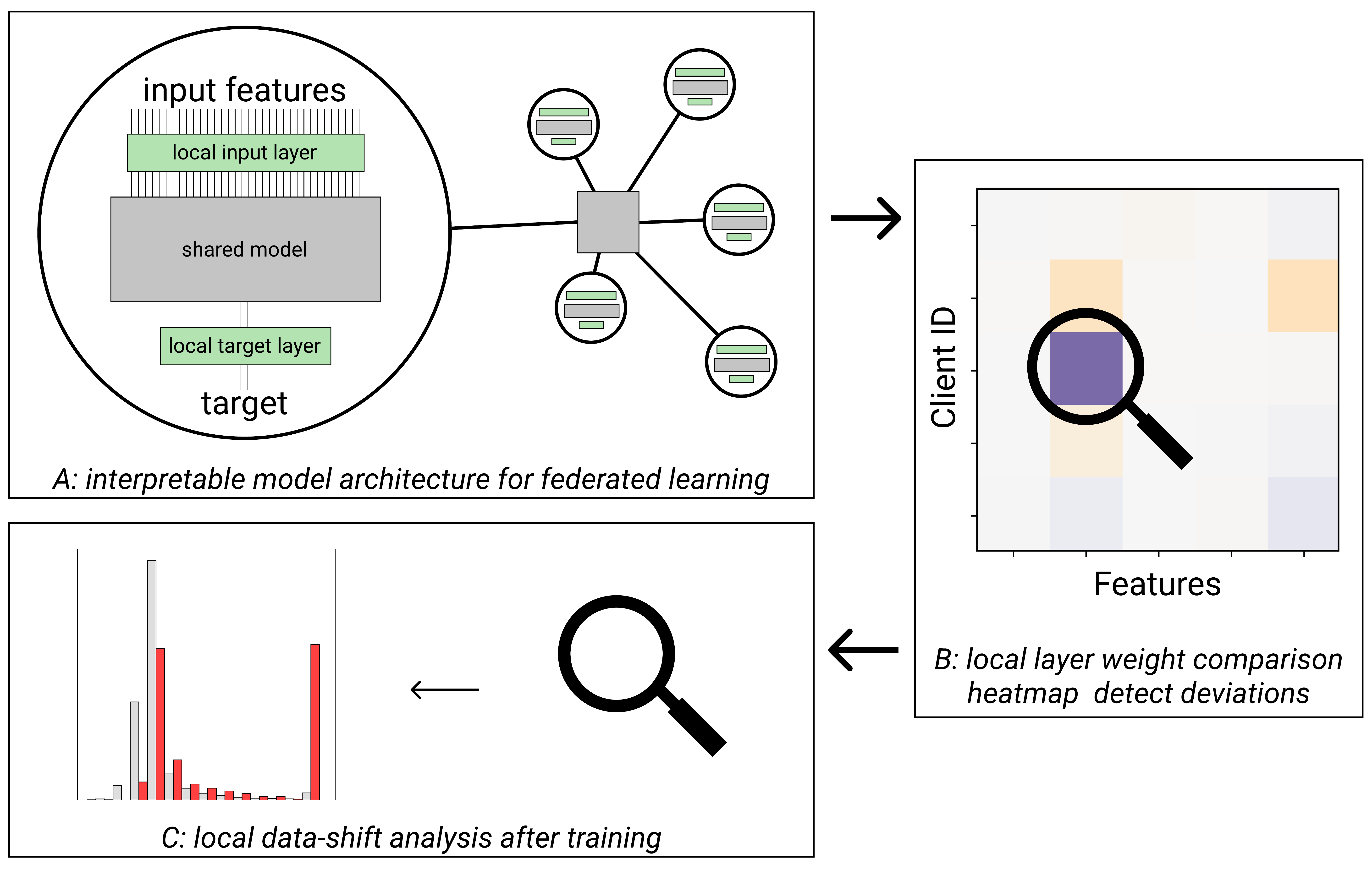}
  \caption{The three phases enabled by \texttt{iFedAvg}. A) a novel model architecture that extracts feature-wise interoperability information deployable in a federated setting B) interpretable outputs allowing practitioners to detect and understand inter-client compatibility issues C) independent private interpretation performed after the training procedure.}
  \label{fig:1_cycle}
\end{figure}

Our proposed method, \texttt{iFedAvg} is designed not only as a personalized federated learning algorithm but as a component of a complete framework. Figure \ref{fig:1_cycle} illustrates this entire workflow, highlighting how \texttt{iFedAvg} enables each step. The following subsections outline the main aspects of the proposed method.

\subsection{Architecture}

Let us denote an existing neural network model which could be shared among all clients as $f_\shared : \R^D \rightarrow \R^K$. This model maps an input vector $\xx \in \R^D$ to a target vector $\yy \in \R^K$. We place no assumptions on the type or complexity of the $f_\shared$'s network as long as it can be trained using gradient-based optimizers and conforms to the definition above. The objective of our work is to devise an extension, allowing for $f_\shared$ to be more interoperable in a transparent fashion.

\texttt{iFedAvg} introduces personalization layers around the shared neural network, $f_\shared$. The combined model is then specified as $f_\out \circ f_\shared \circ f_\inp$, where $\circ$ indicates a composition. To retain the correct dimensionality, the input and output layers are as follows: $f_\inp : \R^D \rightarrow \R^D $ and $f_\out : \R^K \rightarrow \R^K $.

In order to retain interpretability at the feature- and target-level, we propose the layers to simply be an element-wise learned normalization. We assume that numerical values are standardized for each client. Inspired by traditional standardization we explicitly define $f_\inp$, with bias and weight vectors $\bb_\inp, \ww_\inp \in \R^D$ as:

\begin{equation}
f_\inp (\xx) = (\xx + \bb_\inp) \odot \ww_\inp \ ,
\end{equation}

where $\odot$ refers to the element-wise multiplication. This construction does not allow the blending that a traditional \textit{fully-connected} layer permits. $f_\inp$ is initialized to be the identify function, namely setting $\bb_\inp = 0$ and  $\ww_\inp = 1$. Similarly, we can define $f_\out (\yy) = (\yy + \bb_\out) \odot \ww_\out$, with $\bb_\out, \ww_\out \in \R^K$. Furthermore, with only additional $2D + 2K$ parameters, the memory cost is negligible.

At first glance, our approach appears analogous to the \textit{personalization layers} proposed by \citet{arivazhagan2019personalizationlayers}. \texttt{iFedAvg} differs however, in the purposeful placement and heavily restricted design of the layers which enable actionable insights to be extracted. Furthermore, this study does not solely analyze the personalization properties of the method, but focuses on revealing how the federated learning process deals with biases in underlying data as well as offering a means of compensating for them.

\subsection{Training} 
Training \texttt{iFedAvg} does not differ substantially from \texttt{FedAvg}. At each round, each client performs a local update using stochastic gradient descent (SGD) and locally retains the updates on $f_\inp$ and $f_\out$. Updates for all weights of $f_\shared$ are disclosed to the server, which performs the standard \texttt{FedAvg} aggregation step and broadcasts the new shared weights to all clients.

This procedure is significantly more efficient than other related personalization methods such as \texttt{APFL} or \texttt{Per-FedAvg}. From a performance perspective, neither a second backpropagation or a Hessian, respectively, need to be computed. A single iteration of SGD is able to update the entire \textit{combined} model at once. Furthermore, only a single copy of the network needs to be stored, which reduces the storage demands on each client, compared to \texttt{APFL}. 

Intuitively, these layers allow each client to learn a feature-shift as well as target-shift, with respect to the federation. As the central block of the neural network is shared, and therefore identical for each client, the personalized layers can be seen as learning the necessary transformation from the underlying private data to the model of the federation. 

\subsection{Interpretable outputs}

The cornerstone of the interpretability of \texttt{iFedAvg} is the personalized layer design. The layers learn the local shifts necessary to be able to utilize the shared model block of the federation. Each client can, therefore, adjust their \textit{combined} model in a very restricted sense. This restriction-by-design means that each individual value in $f_\inp$ and $f_\out$ is directly interpretable. 

For instance, following the example above, if the age of patients varies from client to client, clients are, at first, indistinguishable after standardization. With our method, however, the necessary personalized age shift can be learned seamlessly throughout the process and can be learned to correctly predict the diagnosis. The magnitude and direction of this shift is, by design, tied directly to the input feature, and provides unparalleled insights about each participant of the federation without sharing any original data.

A crucial distinction must be made between our method and comparing the underlying data distribution a-priori. Exchanging the datasets' summary statistics can be a useful way to diagnose interoperability, but is limited in three critical ways. First, every feature must be analyzed for every client - in practice the needed effort might render this impossible. Second, revealing summary statistics might still be problematic from a data privacy perspective - personalized model weights are more secure in that regard. Third, \texttt{iFedAvg} only learns \textit{necessary} shifts which aid model performance, eliminating the necessity to investigate potential shifts for insignificant features.

The final step of our proposed architecture is a communication of the learned shifts to stakeholders. Large compensations could be an indicator of critical differences in data collection, calibration, missingness-not-at-random or otherwise. A decision on whether this means a client's dataset is incompatible or, with the shift-adjustment, is suitable is left to domain experts. \texttt{iFedAvg} strives to be the best tool to make an informed decision for which clients and features to include in their federation. Significant deviations could then also allow targeted queries between users to better guide collaboration. 

\section{Federated Ebola dataset}
\label{sec:ebola}
The 2014-16 Ebola Virus Disease (EVD) epidemic in West Africa revealed the devastating consequences of inadequate data sharing during public health emergencies. Delays and poorly compatible datasets were held directly responsible for the slow response to the outbreak, ultimately exacerbating the epidemic \citet{ebolageorgetown}. 
In response, the Infectious Disease Data Observatory (IDDO) was established to collate and align the fragmented and poorly interoperable datasets into the largest central repository of Ebola data in the world (i.e., the Ebola Data Platform, EDP) to facilitate coordinated research \citep{ebolaiddo}.

The commendable EDP initiative necessitated a laborious process of acquiring ethical approval as well as devising a common data sharing protocol. While this was ultimately highly successful, it took several years before being made available to researchers in 2019, well after it would have been useful in forming an evidence-based response to the crisis. Thus, the EDP is an ideal real-world use-case for exploring the potential of an interoperability-adjusted federated learning approach with the end goal of enabling secure real-time model sharing in rapidly evolving public health emergencies which suffer from poor response coordination.

The EDP comprises tabular clinical data on 13552 anonymized patients treated at 16 Ebola Treatment Centres (ETCs) between January 2014 and December 2015. The ETCs were scattered across the three main affected countries of Sierra Leone, Guinea and Liberia which differ in language, geography, epidemiology, demographics and treatment protocols. Thus, there is high risk for bias ranging from natural variation (e.g., malaria prevalence) to measurement errors (e.g., different tools/protocols to quantify viral load) and misattribution (e.g., mislabelling or poor standardization) and variable missingness.
The collected data includes both categorical and continuous features such as demographic details (e.g., age, sex, location), clinical signs and symptoms (e.g., fever, coughing, headache), laboratory values (e.g., Ebola test results and quantitation of viral load) and outcomes for each patient (e.g., death vs recovery). Research on such data is often focused on making diagnostic and prognostic models to better allocate limited resources to the most critical patients and improve early case identification \cite{ebolahartley1, ebolahartley2, ebolacolubri, ebolareview}.

However, these studies were performed on single datasets where statistical power is diluted in the small numbers of included patients.   

As a proof of concept, we replicate these studies, by learning diagnostic and prognostic classification tasks (respectively, EVD negative vs EVD positive and survival vs death in EVD positive cases). Here, the ETC site represents a client with its locally collected dataset and we thus explore the potential of \texttt{iFedAvg} to create a robust personalized predictive model while detecting and compensating for known interoperability issues between sites.

The ethical framework and anonymization protocols are published on the IDDO website \citep{ebolaiddo}. We provide further details such as sample sizes and class imbalance in Appendix \ref{app:ebola}.

\section{Experimental design}
\label{sec:experimentalsetup}
\textbf{Evaluation}. To fairly evaluate \texttt{iFedAvg}, we compare it to several state-of-the-art methods which have similar aims, and explore their limitations in a standardized realistic setting. Specifically, each client can choose to not partake in the federation, and simply train a local model. Likewise, vanilla federated averaging, \texttt{FedAvg} is a valuable benchmark, as well as a more sophisticated personalized federated algorithm such as Adaptive Personalized Federated Learning, \texttt{APFL}, proposed by \citet{deng2020APFL}. An interesting non-personalized baseline is a single model trained on a centralized concatenation of all clients' datasets (called in the following the Centralized method). While this is not a realistic scenario, it is currently the yardstick of many large scale data sharing efforts such as IDDO, which makes it a particularly appropriate benchmark for the Ebola dataset. 

Four datasets will serve as the foundation of our experiments:
\begin{itemize}
    \item \textbf{Ebola Prognosis:} Predicting the survival of EVD-positive patients; \citep{ebolaiddo}.
    \item \textbf{Ebola Diagnosis:} Predicting whether a patient triaged as "suspect" has EVD;  \citep{ebolaiddo}.
    \item \textbf{Vehicle Classification with Sensor Network (VSN):} Classifying the type of vehicle based on a network of 23 acoustic and seismic sensors; \citep{duarte2004VSN}.
    \item \textbf{Human Activity Recognition (HAR):} Classifying the type of activity performed by 30 human subjects according to readings from body sensors; \citep{anguita2013HAR}.
\end{itemize}
Every method outlined above is trained on each dataset, with each client retaining 33\% or minimally 100 samples of its local data as a hold-out test. Performance metrics are computed locally on this hold-out set, retaining an understanding of personalized performance. 

\textbf{Preprocessing \& missing values}. The experimental setting assumes semantic interoperability, meaning all client datasets were aligned in feature nomenclature. All numerical values are standardized to mean 0 and standard deviation 1. In order to include missingness as an assessed feature of interoperability (i.e., to detect whether the bias is due to non-random missing values), missing continuous values were filled with 0s, binary features with the value 0.5 following standardization.

\textbf{Architecture}. For every method, an MLP model with identical architecture is used, only modifying the training regime for each algorithm. In the main experimental results we only enable $f_\inp$ to investigate feature-shifts. For results and a thorough discussion of \texttt{iFedAvg} with $f_\out$, we refer to Appendix \ref{app:target}. 

\textbf{Training}. Each round, every client performs one training epoch on the local training data, with a weighted loss to account for class imbalance. For the shared part of the model, $f_\shared$, uniform weighting of the updates across clients is performed. Every experiment is conducted on the same 5 random seeds, leading to identical initialization and train-test splitting. The code, implemented in PyTorch \citep{paszke2019pytorch}, replicating the results on the public datasets, are available in a public code repository.\footnote{\href{https://github.com/davidroschewitz/ifedavg}{\texttt{github.com/davidroschewitz/ifedavg}}} Further details on the hyperparameters, pre-processing and the model architecture can be found in Appendix \ref{app:params_experiment}. 

\section{Results}
\label{sec:results}

\subsection{Performance}
From the perspective of a client choosing whether to participate in the federation, predictive performance is critical. In particular, attaining a collaborative model \textit{worse} than a locally trained one virtually rules out participation. Therefore, two aspects are interesting: the average metric across all clients and the worst-performing client in the federation.

Holistically, \texttt{iFedAvg} shows competitive performance compared with a state-of-the-art benchmark in distributed personalized learning algorithm, \texttt{APFL}, and even outperforms both \texttt{APFL}, local and centralized training in several instances. As anticipated, in the case of heterogeneous or non-IID datasets, such as Ebola diagnosis and VSN, \texttt{iFedAvg} vastly outperforms \texttt{FedAvg}. Personalization appears important to adapt to these settings. Table \ref{tab:mean} shows the average F1 score for each dataset and algorithm.

\begin{table}
  \caption{Mean (across all clients) performance (F1 score)}
  \label{tab:mean}
  \centering
  \begin{tabular}{lccccc}
    \toprule
    & \multicolumn{4}{c}{Method}                   \\
    \cmidrule(r){2-6}
    Dataset & \texttt{iFedAvg} (ours) & \texttt{APFL} & \texttt{FedAvg} & Local & Centralized\\
    \midrule
    Ebola Prognosis &	0.669 & 0.653 & 0.670 & 0.662 & \textbf{0.673} \\
    Ebola Diagnosis &	\textbf{0.867} & 0.828 & 0.773 & 0.844 & 0.861 \\
    Vehicle Sensor Network &	0.928 &	0.935 &	0.871 &	0.939 &	\textbf{0.943} \\
    Human Activity Recognition &	\textbf{0.994} & 0.992 & 0.967 & 0.993 & 0.988 \\
    \bottomrule
  \end{tabular}
\end{table}

Analyzing the worst-performing client highlights the low tail of the performance distribution. In Table \ref{tab:worst} we can observe that \texttt{FedAvg} performs especially poorly in this worst case. Intuitively, the client with a significantly shifted data distribution is \textit{forced} to share the same global model, leading to inferior performance. This experiment highlights that \texttt{iFedAvg} is an effective method to personalize collaborative learning and is especially robust for the worst performing client in the federation. Additional tables and visualizations highlighting the full distribution of client performances, seed variation and additional metrics can be found in Appendix \ref{app:performance_results}. 

\begin{table}
  \caption{Worst-performing client in federation (F1 score)}
  \label{tab:worst}
  \centering
  \begin{tabular}{lccccc}
    \toprule
    & \multicolumn{4}{c}{Method}                   \\
    \cmidrule(r){2-6}
    Dataset & \texttt{iFedAvg} (ours) & \texttt{APFL} & \texttt{FedAvg} & Local & Centralized\\
    \midrule
    Ebola Prognosis & \textbf{0.581} & 0.546 & 0.575 & 0.560 & 0.541 \\
    Ebola Diagnosis & \textbf{0.790} & 0.662 & 0.452 & 0.654 & 0.733 \\
    Vehicle Sensor Network & 0.874 & 0.874 & 0.398 & \textbf{0.884} & 0.834 \\
    Human Activity Recognition & 0.950 & 0.972 & 0.909 & \textbf{0.974} & 0.955 \\
    \bottomrule
  \end{tabular}
\end{table}

\subsection{Interpretability of \texttt{iFedAvg}}
One of the key improvements our method provides is that each locally personalized layer is directly interpretable. While the absolute value of each weight does not itself imply an exact relationship with the underlying data, the magnitude, direction and, most importantly, the comparison with other clients' local weights, provides unparalleled insights into the interoperabiltiy of the datasets in the federation. 

For the most valuable interpretable results, all clients are assumed to be willing to share the weights of their locally trained personalized layers, $\bb_\inp$ and $\ww_\inp$ for our experiments. In practice, this is a reasonable expectation, as raw data is not revealed and the insights might be critical for the clients.

\begin{figure}
  \centering
  \includegraphics[width=0.99\linewidth]{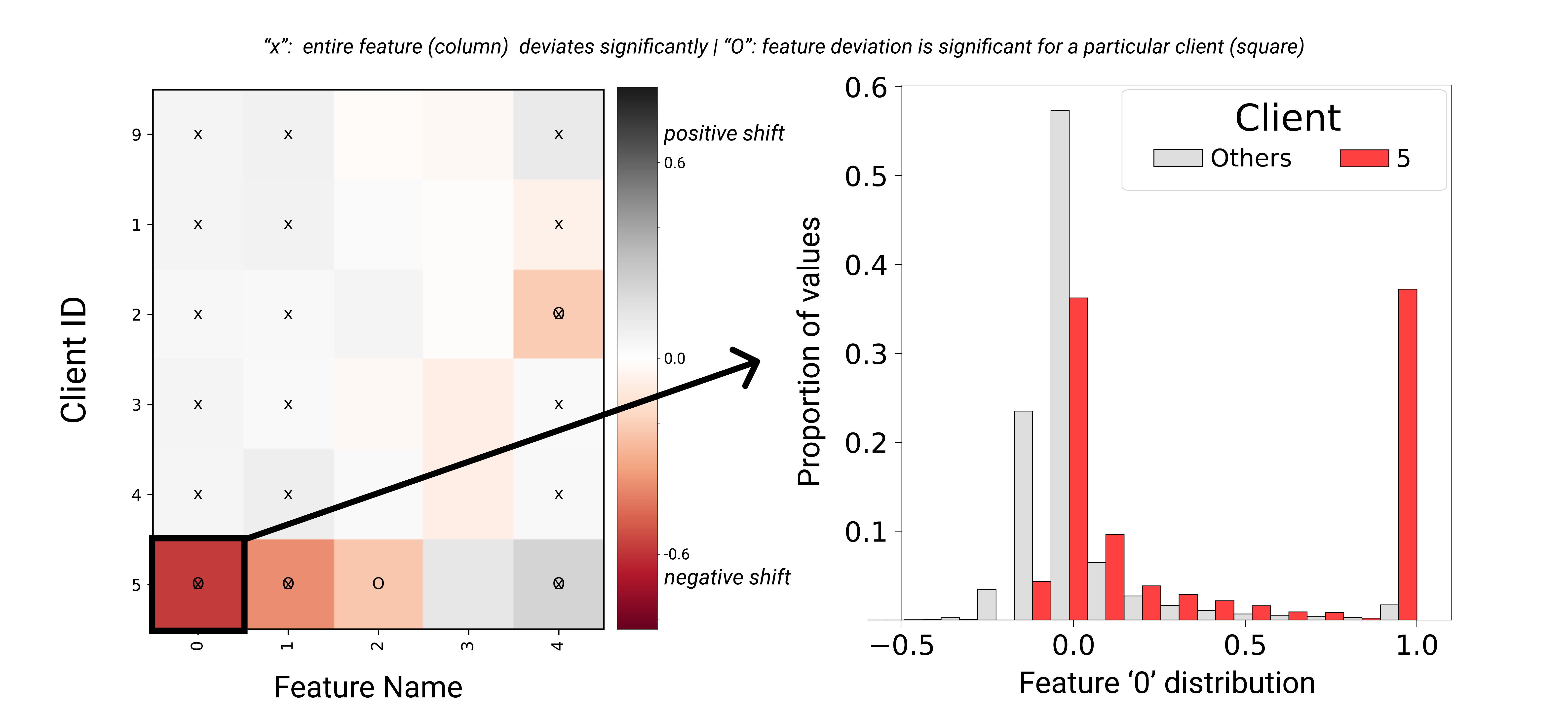}
  \caption{Input bias ($\bb_\inp$) heatmap across clients for the VSN dataset. A negative shift is detected for sensor number 5 in feature `0' (left), which can be confirmed in the underlying distribution (right).}
  \label{fig:results_example_vehicle}
\end{figure}

In order to visually inspect the personalized shifts, we combine the weights into a heatmap for every feature and client to create a 2D matrix for each bias and weight. In the heatmap, we consider two types of shifts as \textit{significant}. First, when for a single feature, a client's local weight differs from the mean by more than 2 standard deviations (SD). Second, if the average SD of a feature in question differs by more than 2 SD from the average SD across \textit{all features}. 
We show an example of such a heatmap on the VSN dataset in Figure \ref{fig:results_example_vehicle}, with the corresponding shift in the underlying data. Noteworthy is that our model detected the shift and indicated its direction without access to data of other clients, unlike the diagnosis histogram.

\begin{figure}
  \centering
  \includegraphics[width=0.99\linewidth]{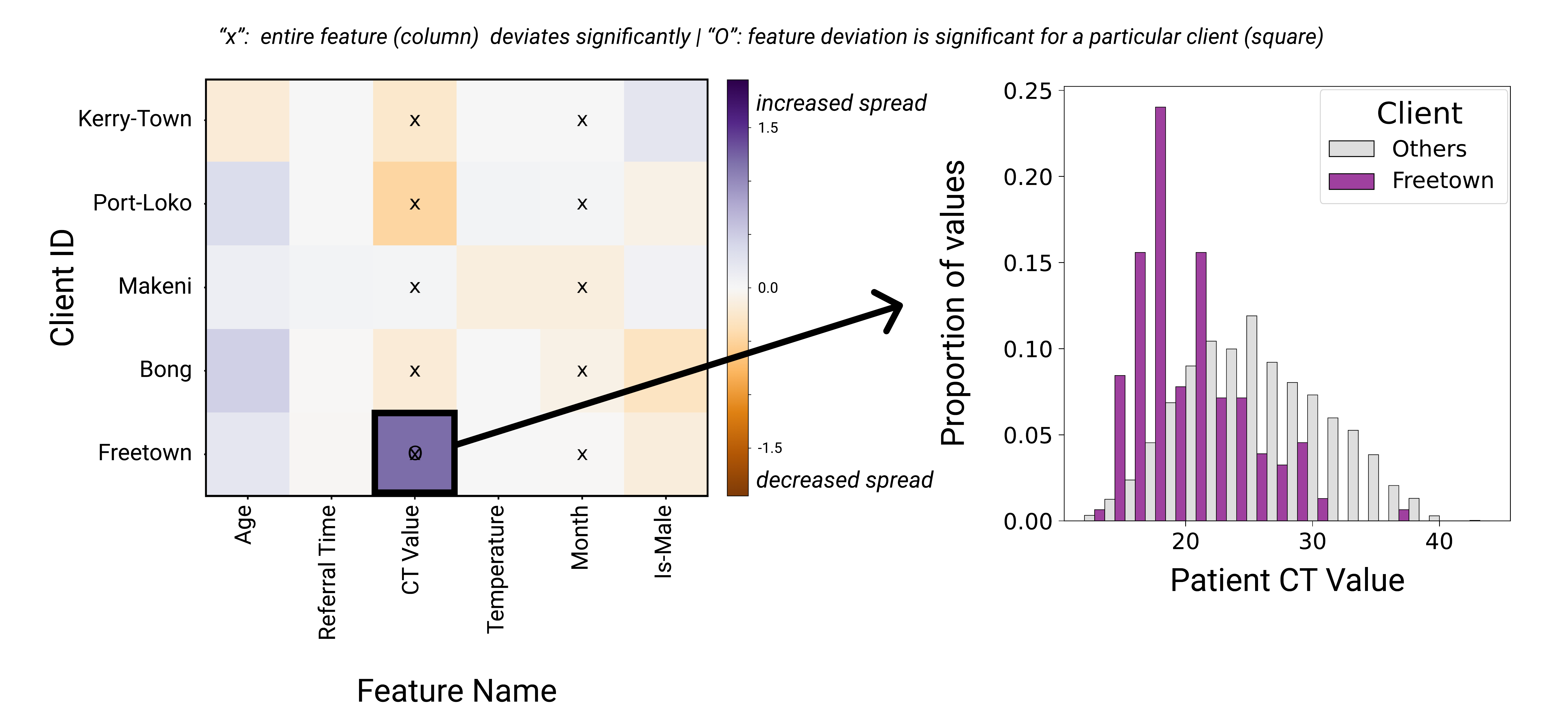}
  \caption{Heatmap of local input weights ($\ww_\inp$) for Ebola patient prognosis (left) and underlying distribution of CT values for ETC `Freetown' compared to all remaining clients (right).}
  \label{fig:prognosis}
\end{figure}

\begin{figure}
  \centering
  \includegraphics[width=0.9\linewidth]{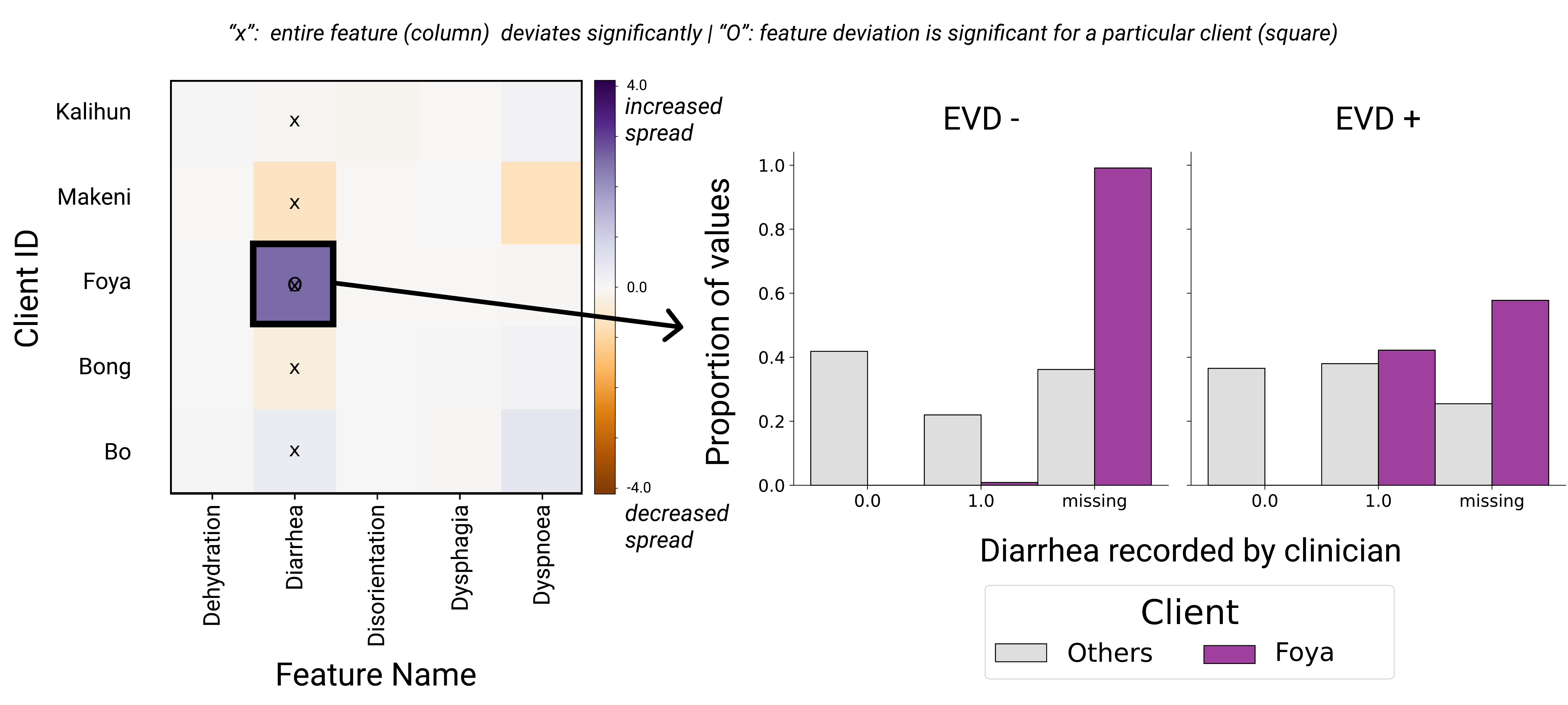}
  \caption{Local input weights heatmap ($\ww_\inp$) for EVD diagnosis (left) and proportion of recorded diarrhea in the patient, split by final diagnosis for ETC `Foya' compared to all remaining clients (right).}
  \label{fig:diagnosis}
\end{figure}

We further highlight the following examples of real-world shifts detected by our method:
\begin{itemize}
    \item \textbf{Ebola Prognosis:} In Figure \ref{fig:prognosis}, we show the heatmap for patient prognosis. Every client significantly modifies the \textit{CT Value} feature compared to other features. We can observe ETC Freetown's distribution being more concentrated, and is therefore scaled by \texttt{iFedAvg} to increase compatibility with the shared model.
    \item \textbf{Ebola Diagnosis:} Similarly, we can observe effects for categorical differences. Figure \ref{fig:diagnosis} highlights that for ETC Foya, having the symptom of diarrhea is a clear indicator of positive Ebola infection (right in Figure). However this is confounded by systematic missingness for the diarrhea feature amongst EVD negative patients. \texttt{iFedAvg} corrects and identifies this difference in data collection without ever sharing any underlying data (left in Figure).
\end{itemize}

In summary, we have shown that \texttt{iFedAvg} is able to detect and correct various types of data shifts across clients in a collaborative learning setting that would ordinarily cause hidden bias and confounding. For highly non-IID datasets, we find large shift compensations (i.e., many significant values in the heatmap visualizations). This behavior is expected, but could be an indication of a non negligible false-positive rate. Given the objective of highlighting \textit{potentially problematic} datasets and features, we believe this tradeoff is acceptable. 

Interestingly, some of the personalized shifts detectable were indicative of data collection bias. Without a mechanism to flag such a shift, a personalized model trained by other methods would learn biased insights from data missing not-at-random (MNAR) and create poorly generalizable predictions. For example, let us recall the data collection of diarrhea presence for ETC Foya (Figure \ref{fig:diagnosis}). A locally trained or strongly personalized federated model would achieve excellent predictive performance as patients with diarrhea appear to be correlated with Ebola infection. This pattern does not hold overall, and if falsely learned could lead to misdiagnosis without detection. In instances precisely like this, \texttt{iFedAvg} proves invaluable in detecting local \textit{model biases}. 

For various additional experimental results on the benchmark datasets as well as artificially introduced shifts and mutations confirming the efficacy of our method we refer to Appendix \ref{app:interpretability_results}.

\section{Conclusion}
We study interpretability and inter-client interoperability in federated learning and show how a feature-wise personalized learning approach addresses this challenge. Our framework, \texttt{iFedAvg}, proposes a simple extension to federated averaging that creates interpretable data-interoperabilty between clients by personalizing models. The learned weights ultimately reveal novel insights about the federated learning process as a whole.

On real-world datasets, \texttt{iFedAvg} is competitive with state-of-the-art personalized learning methods in terms of performance. The method vastly outperforms \texttt{FedAvg} or centralized learning on poorly performing clients with significant data shifts. More critically, significant feature-wise shifts in the underlying client datasets are correctly detected and compensated for. Not only does this provide targeted guidance to practitioners interpreting the results, but it can aid assessments of the overall compatibility of a client with the federation. While some shifts might not be harmful to the model, overcompensating for local data biases can be detrimental. \texttt{iFedAvg} visualizes these shifts, therefore generating the necessary transparency to best verify the reliability of model. 

We leave as future research the extension of our approach to other types of data (such as images) or learning a feature alignment mapping. Furthermore, studying \texttt{iFedAvg} with a large number of clients and partial participation would allow more widespread adoption in the future.

In conclusion, \texttt{iFedAvg} offers novel insights into the federated learning process of tabular datasets. It leverages these insights not only for interpretabilty, but to build personalized models that are further adjusted for the usually hidden interoperability issues between clients in a federation. These unique extensions of the federated learning process come at a negligible computational overhead and thus \texttt{iFedAvg} is a promising approach for real world collaborative learning.

\subsection*{Broader impact}
This work is specifically designed to provide more guarantees of interoperability in federated learning and therefore incentivize collaboration, especially in fields with sensitive data and a high risk of collection bias. Equally we could disincentivize interoperable data collection by creating a shortcut that may undermine standardization efforts.
Interoperability is a pillar of the FAIR Guiding Principles for ethical scientific data stewardship, and this critical issue was highlighted as a key barrier by a WHO-commissioned investigation into the massive failings of the centralized Ebola response~\cite{fair2016}. Our work is specifically motivated by this use-case and is appropriately evaluated on a unique dataset collated from the largest number of distributed data collection sites during the notoriously poorly coordinated 2014-16 West African Ebola epidemic. By extrapolation, \texttt{iFedAvg} can be seen as a first step to facilitating interpretable interoperability in collaborative analyses. Basing the architecture on a distributed learning system, we also attempt to address the issue of local data ownership and data privacy compared to centralized approaches. While the insights shared do not reveal sample level data, the client-level aggregate could be considered sensitive and may be abused to discriminate against clients. In this instance, concealing the identity of the client could be considered as a mitigation strategy.
Finally our simplified approach with low computational overhead makes this an accessible method for low-resource settings to build collaborative models whilst better securing patient privacy, intellectual property and statistical robustness to biases between collaborating datasets. 

\section*{Acknowledgements}
The authors would like to acknowledge all the patients whose data was used in this study. This work was inspired by the challenges of sensitive data management in health emergencies and using the Ebola dataset provided critical validation and context. The data was provided by the Ebola Data Platform hosted by the Infectious Diseases Data Observatory (IDDO), and the data contributors, who had no role in the production of these research outputs. The contributors are: Alliance for International Medical Action (ALIMA), International Medical Corps (IMC), Institute of Tropical Medicine Antwerp (ITM), Médecins Sans Frontières (MSF), Oxford University and Save the Children (SCI). We also thank Aiyu Liu for his work in preparing the Ebola dataset for analysis and Mélanie Bernhardt for her critical and insightful thoughts and generous support.

\bibliographystyle{abbrvnat}
\bibliography{references}

%\end{document} %MJ: uncomment and compile just once (then the internal links still work) or cut in mac preview
\newpage
\appendix

\section{Appendix}

\subsection{Supplementary performance results}
\label{app:performance_results}

In addition to the main results in the paper, we provide the full distribution of performance metrics across clients as well as balanced accuracy and ROC AUC scores for each method. In the client performance distribution plots, the red error bars show the standard deviation (SD) of the median score of each random seed. We notice no particular pattern or single method with a distinctive behavior with regards to seed. The following two sections evaluate this experiment by F1 score and AUROC and . 

\subsubsection{F1 Score}
\begin{figure}[H]
  \centering
  \includegraphics[width=0.9\linewidth]{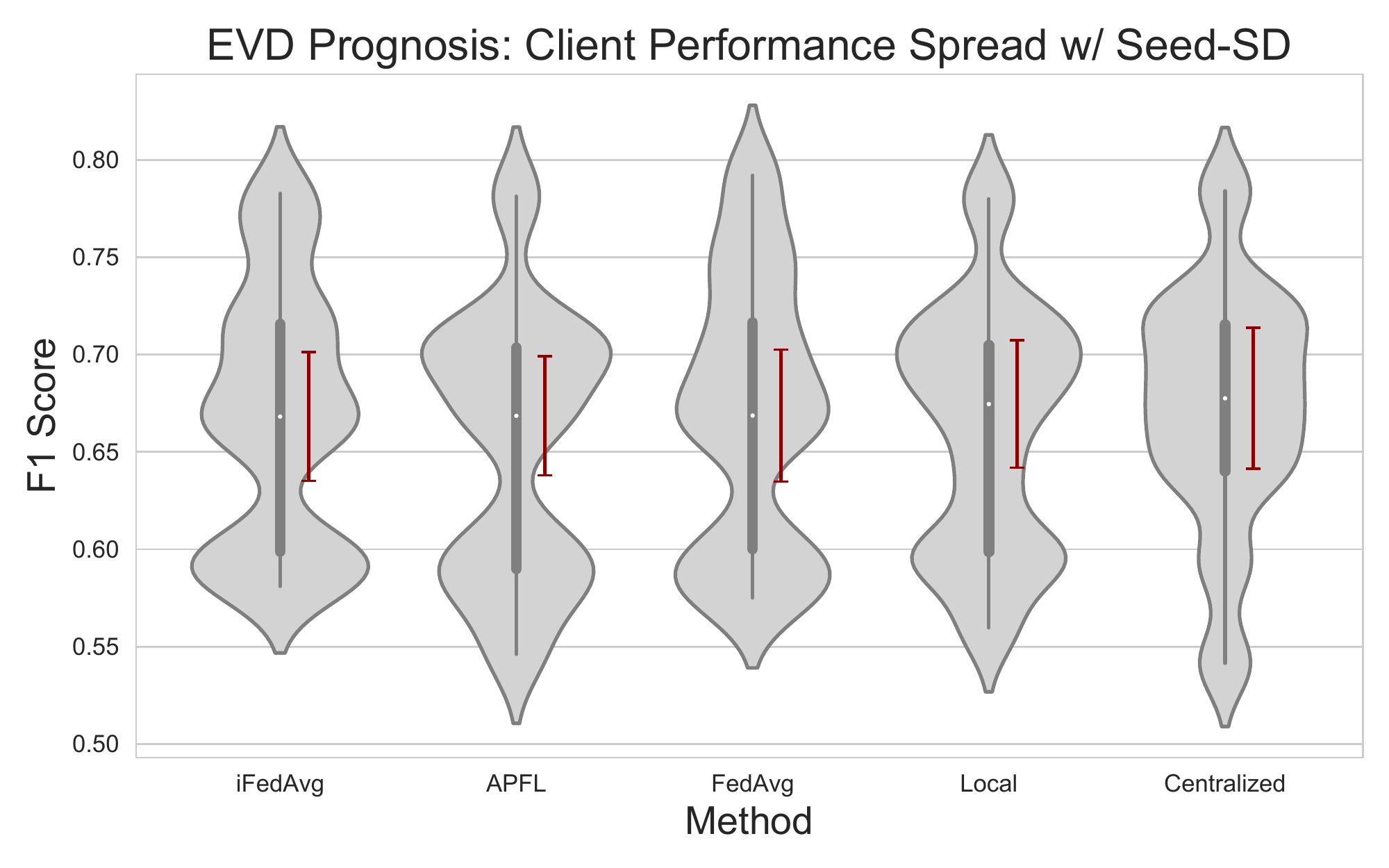}
  \caption{Distribution of client performances (F1 score) for the Ebola Prognosis dataset. Red error bars show the standard deviation (SD) of the median score of each random seed.}
  \label{fig:prog-spread-f1}
\end{figure}

\begin{figure}
  \centering
  \includegraphics[width=0.9\linewidth]{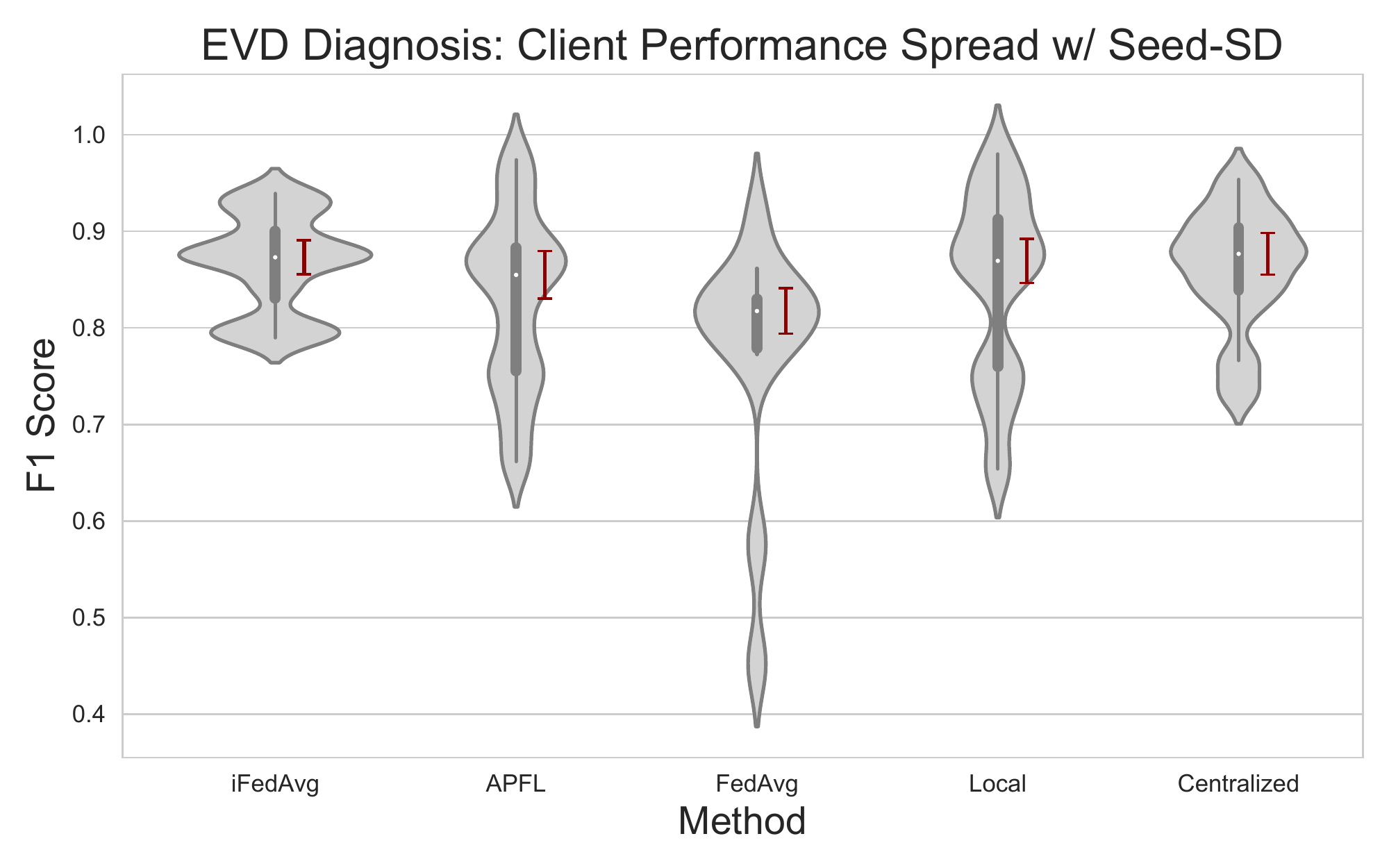}
  \caption{Distribution of client performances (F1 score) for the Ebola Diagnosis dataset. Red error bars show the standard deviation (SD) of the median score of each random seed.}
  \label{fig:diag-spread-f1}
\end{figure}

\begin{figure}
  \centering
  \includegraphics[width=0.9\linewidth]{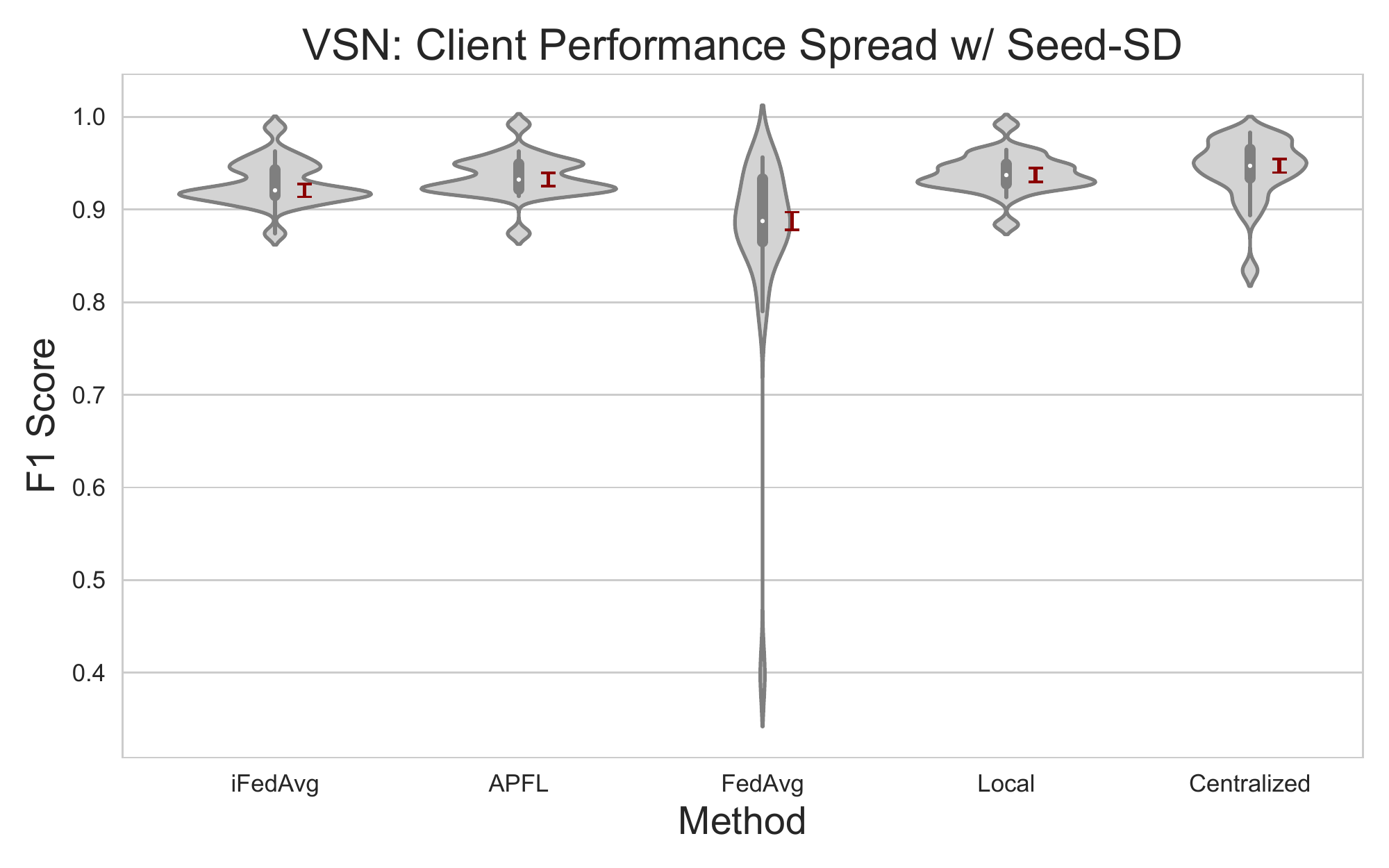}
  \caption{Distribution of client performances (F1 score) for the Vehicle Sensor Network dataset.Red error bars show the standard deviation (SD) of the median score of each random seed.}
  \label{fig:vehicle-spread-f1}
\end{figure}

\begin{figure}
  \centering
  \includegraphics[width=0.9\linewidth]{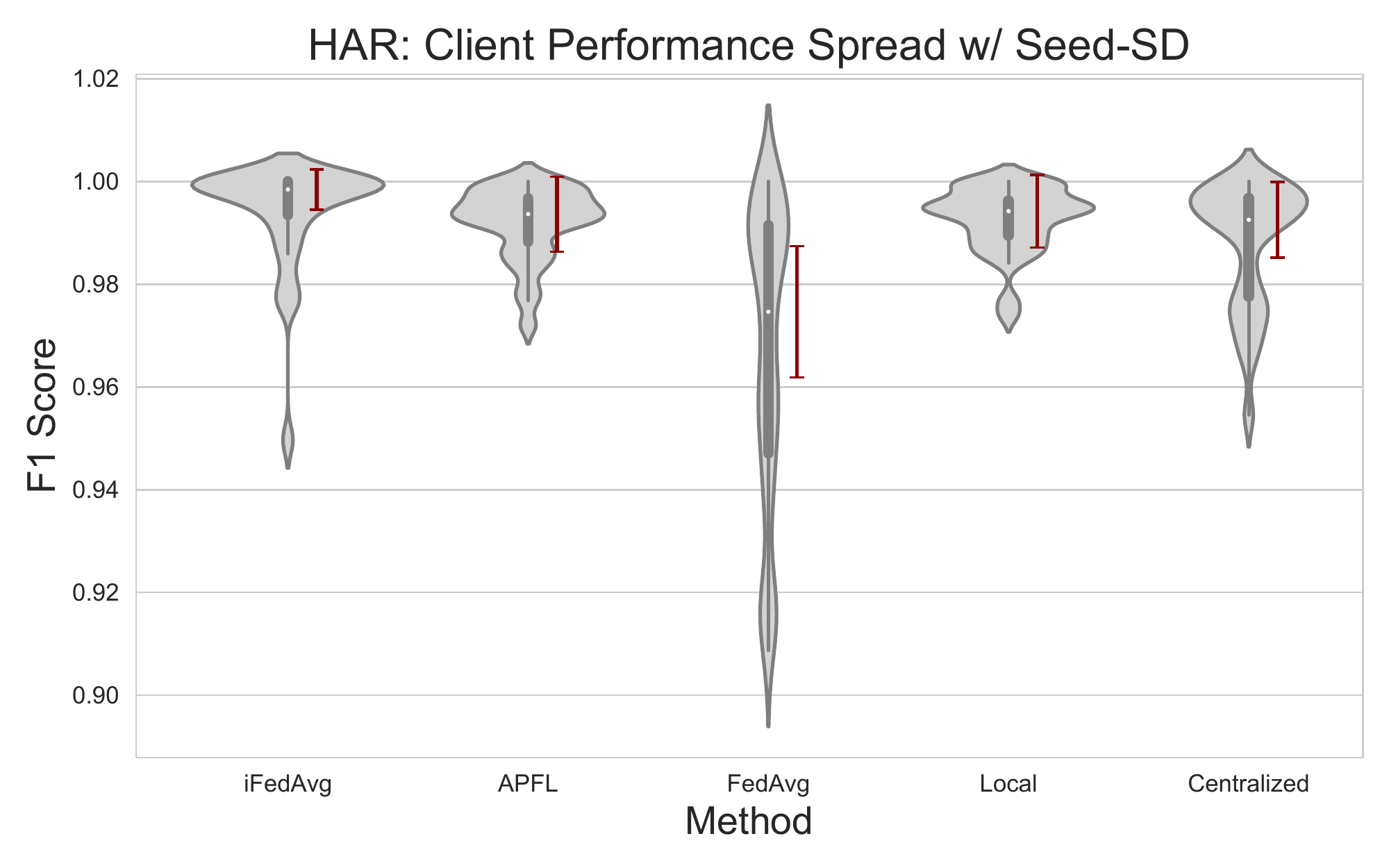}
  \caption{Distribution of client performances (F1 score) for the Human Activity Recognition dataset. Red error bars show the standard deviation (SD) of the median score of each random seed.}
  \label{fig:har-spread-f1}
\end{figure}

Tables \ref{tab:mean} and \ref{tab:worst} (in the main text) show mean and worst-client performance on all datasets. Here, we show the distribution of client performances in violin plots for each dataset (EVD Prognosis: Figure \ref{fig:prog-spread-f1}, EVD Diagnosis: Figure \ref{fig:diag-spread-f1}, VSN: Figure \ref{fig:vehicle-spread-f1}, HAR: Figure \ref{fig:har-spread-f1}). 

It is particularly apparent in these visualizations that \texttt{iFedAvg} is robust to \textit{poorly} performing clients. For every dataset our method outperforms \texttt{FedAvg}, and in most instances even outperforms \texttt{APFL}, Local or Centralized training. Furthermore, for the HAR dataset, while \texttt{iFedAvg} has a relatively low performing client, the overall distribution is skewed towards $1.0$, indicating good overall performance. For this dataset, the seed-SD is also noticeably lower for our method compared to the benchmarks.


\newpage
\subsubsection{ROC AUC}
An additional metric which can be used to measure the performance of a classification model is the area under the curve of the receiver operating characteristic. We evaluate our method in the same fashion as previously; analyzing both the average performance across clients as well as the worst performing client in the federation. These results are displayed in Table \ref{tab:rocauc-mean} and \ref{tab:rocauc-worst}.

These graphs corroborate the findings in the main text, and highlight that the strong performance of \texttt{iFedAvg} is independent of the chosen metric. Interestingly, the gap between our method and vanilla federated averaging shrinks marginally, which can be explained by the sensitivity of F1 score to individual incorrect samples.

\begin{table}
  \caption{Mean (across all clients) performance (ROC AUC)}
  \label{tab:rocauc-mean}
  \centering
  \begin{tabular}{lccccc}
    \toprule
    & \multicolumn{4}{c}{Method}                   \\
    \cmidrule(r){2-6}
    Dataset & \texttt{iFedAvg} (ours) & \texttt{APFL} & \texttt{FedAvg} & Local & Centralized\\
    \midrule
    Ebola Prognosis & 0.725 & 0.704 & \textbf{0.726} & 0.708 & 0.74 \\
    Ebola Diagnosis & \textbf{0.909} & 0.860 & 0.879 & 0.870 & 0.901 \\
    Vehicle Sensor Network & 0.975 & 0.978 & 0.921 & 0.982 & \textbf{0.986} \\
    Human Activity Recognition & 1.00 & 1.00 & 0.999 & 1.00 & 1.00 \\
    \bottomrule
  \end{tabular}
\end{table}

\begin{table}
  \caption{Worst-performing client in federation (ROC AUC)}
  \label{tab:rocauc-worst}
  \centering
  \begin{tabular}{lccccc}
    \toprule
    & \multicolumn{4}{c}{Method}                   \\
    \cmidrule(r){2-6}
    Dataset & \texttt{iFedAvg} (ours) & \texttt{APFL} & \texttt{FedAvg} & Local & Centralized\\
    \midrule
    Ebola Prognosis & \textbf{0.628} & 0.573 & 0.608 & 0.567 & 0.581 \\
    Ebola Diagnosis & \textbf{0.799} & 0.694 & 0.770 & 0.689 & 0.740 \\
    Vehicle Sensor Network & 0.930 & 0.936 & 0.124 & \textbf{0.950} & 0.935 \\
    Human Activity Recognition & 0.996 & 0.997 & 0.991 & 0.997 & \textbf{0.998} \\
    \bottomrule
  \end{tabular}
\end{table}

\begin{figure}
  \centering
  \includegraphics[width=0.9\linewidth]{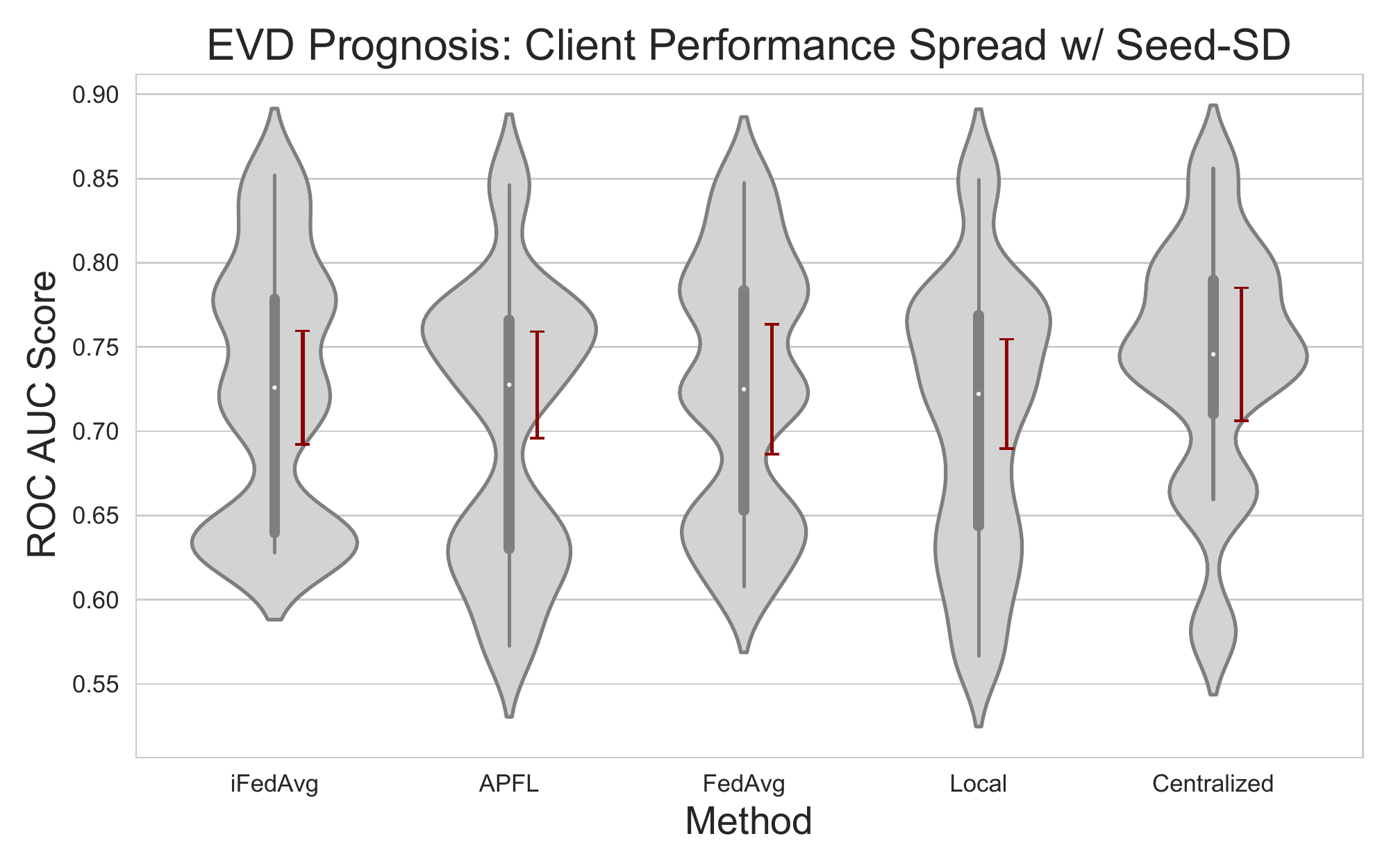}
  \caption{Distribution of client performances (ROC AUC) for the Ebola Prognosis dataset.}
  \label{fig:prog-spread-rocauc}
\end{figure}

\begin{figure}
  \centering
  \includegraphics[width=0.9\linewidth]{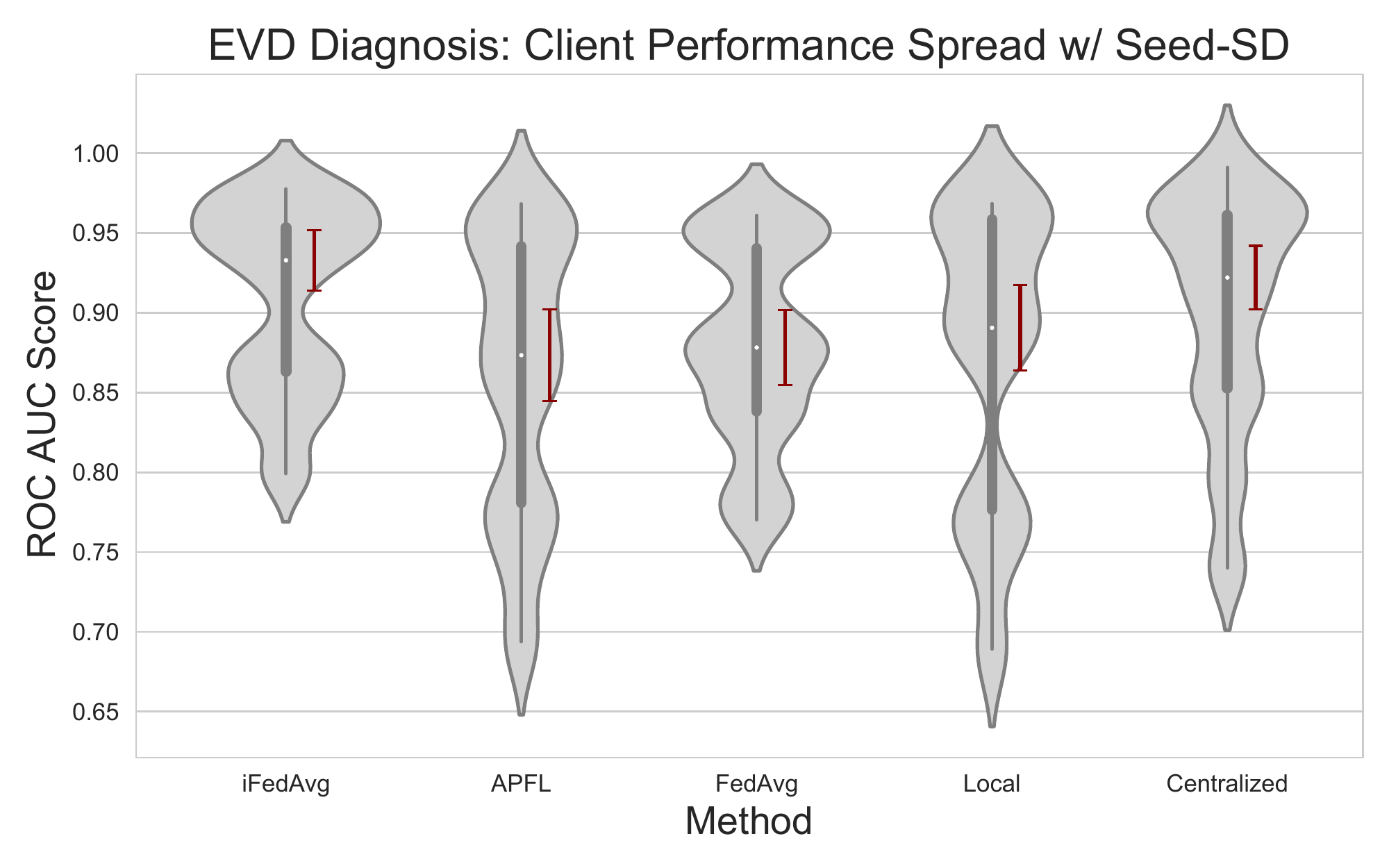}
  \caption{Distribution of client performances (ROC AUC) for the Ebola Diagnosis dataset.}
  \label{fig:diag-spread-rocauc}
\end{figure}

\begin{figure}
  \centering
  \includegraphics[width=0.9\linewidth]{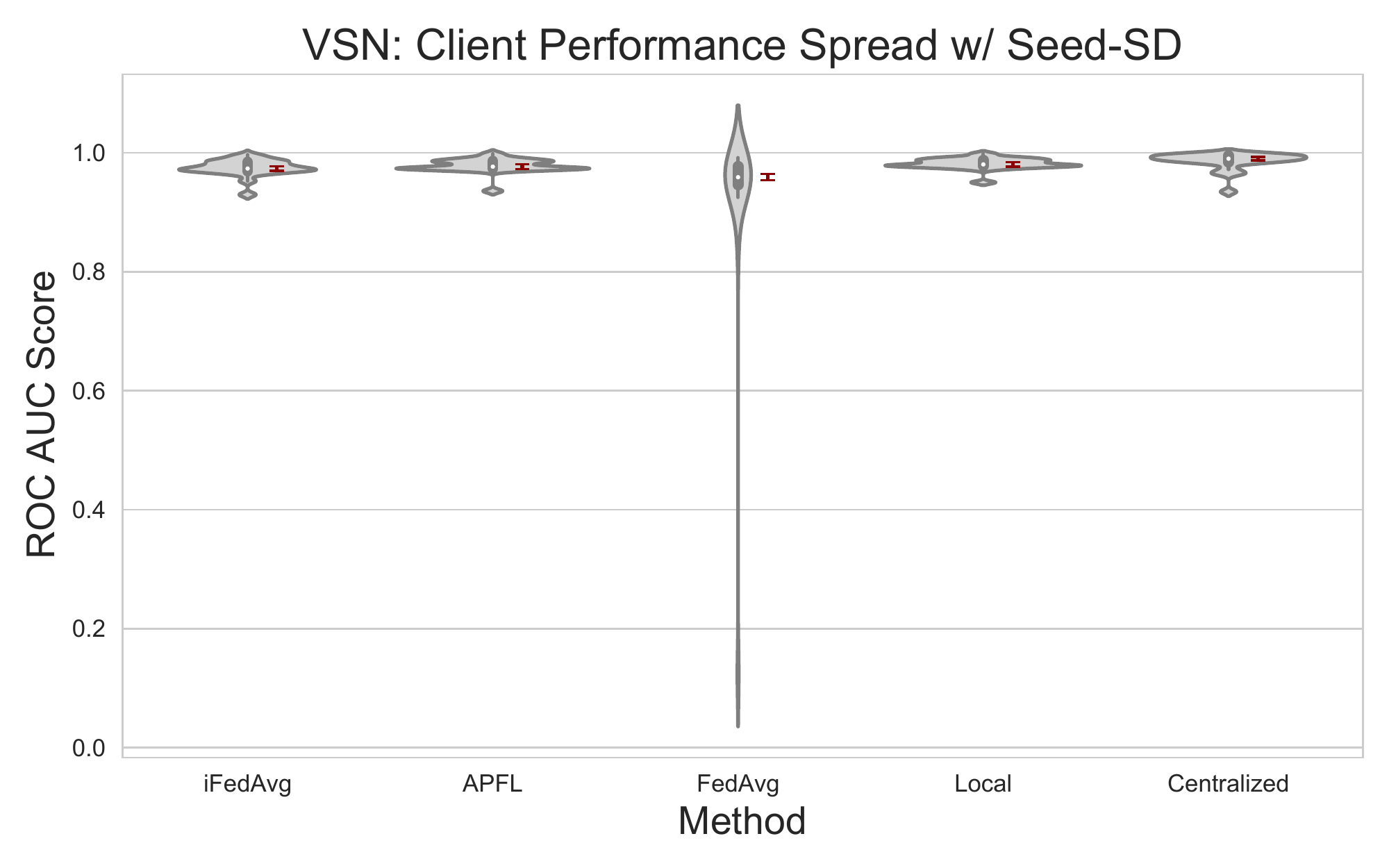}
  \caption{Distribution of client performances (ROC AUC) for the Vehicle Sensor Network dataset.}
  \label{fig:vehicle-spread-rocauc}
\end{figure}

\begin{figure}
  \centering
  \includegraphics[width=0.9\linewidth]{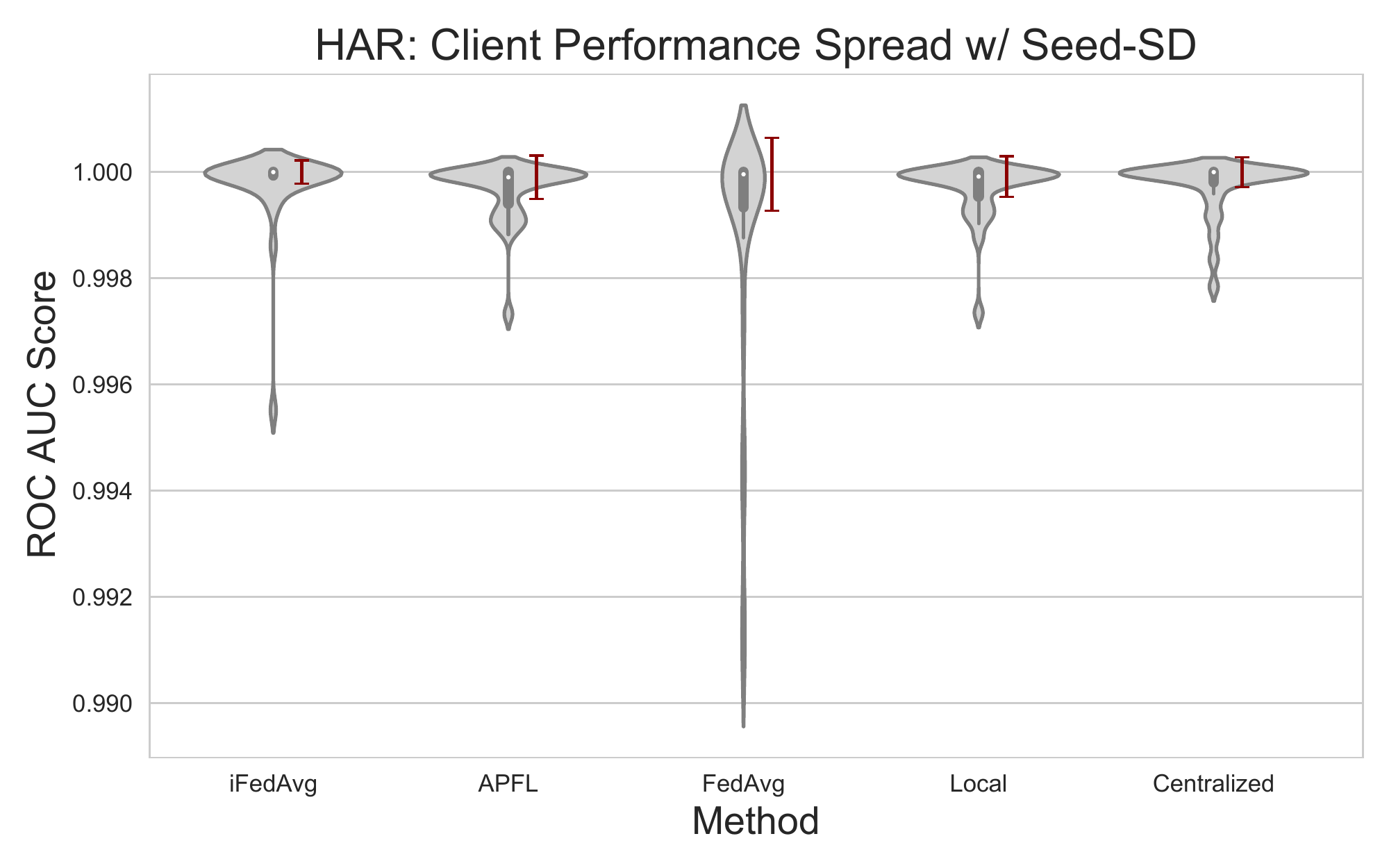}
  \caption{Distribution of client performances (ROC AUC) for the Human Activity Recognition dataset.}
  \label{fig:har-spread-rocauc}
\end{figure}

Similarly to the F1 score metric, we present all distributions of ROC AUC performance as violin plots (EVD Prognosis: Figure \ref{fig:prog-spread-rocauc}, EVD Diagnosis: Figure\ref{fig:diag-spread-rocauc}, VSN: Figure \ref{fig:vehicle-spread-rocauc}, HAR: Figure \ref{fig:har-spread-rocauc}). One instance that stands out is the relatively poor performance of \texttt{APFL} for EVD Prognosis and Diagnosis, as the personalized method does not manage to outperform vanilla federated averaging. We suspect that \texttt{APFL} is overfitting due to the strongly heterogeneous nature of the datasets, which can be observed in its similarity to Local training performance.


In conclusion, \texttt{iFedAvg} shows impressive performance in various settings and measured by different metrics. Especially on datasets which benefit from personalization due to one or multiple outlier clients, our method is particularly effective compared with vanilla federated averaging.

\newpage
\subsection{Supplementary interpretability results}
\label{app:interpretability_results}
In this section, we show additional interpretability results of \texttt{iFedAvg}, with the objective of highlighting the different types of shifts detected for various datasets. The list below serves as a reference to the corresponding figures. 

\begin{itemize}
    \item \textbf{Human Activity Recognition - underlying shifts:} In order to determine the directionality of detected shifts, we highlight two examples that are shown for \texttt{iFedAvg}. For feature 77, we notice that all shifts are significant (`x' in the column), but two local weights for clients 14 and 15 are additionally significant (`O'). We can see that the positive and negative shifts in bias ($\bb_\inp$) in grey and red, respectively, are noticeable in the underlying data. The histograms clearly show that the values of client 15 are skewed towards $-1.0$ whereas for client 14 they are skewed towards $1.0$. This example demonstrates that the learned local layers can indicate directionality correctly. Figure \ref{fig:har_bias} shows the bias heatmap as well as the histograms of both clients of interest.
    \begin{figure}
      \centering
      \includegraphics[width=0.99\linewidth]{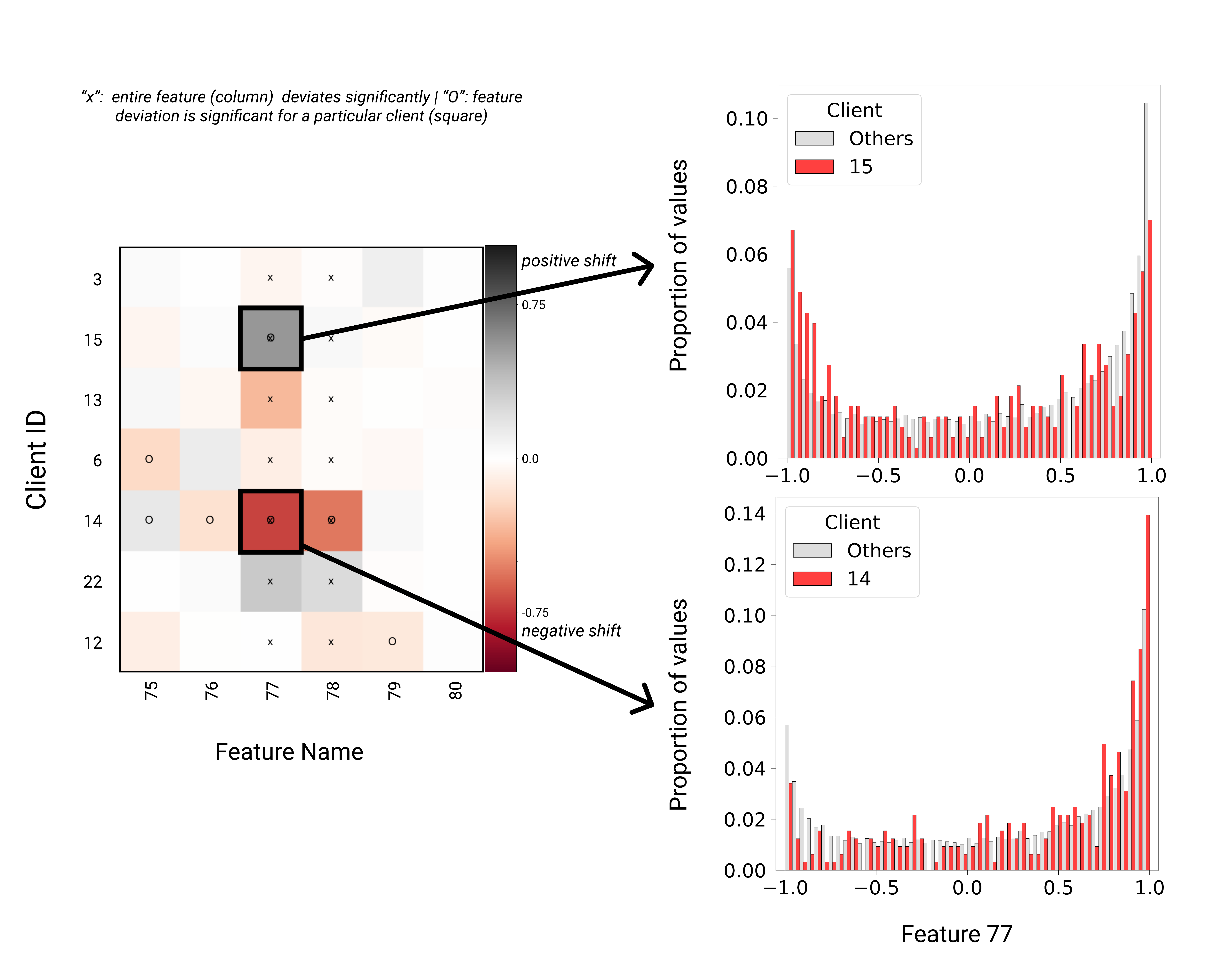}
      \caption{Heatmap of the biases ($\bb_\inp$) for the Human Activity Recognition dataset (left) and histograms of feature 77 for clients 15 and 14 (right).}
      \label{fig:har_bias}
    \end{figure}
    
    \item \textbf{Human Activity Recognition - feature spread:} Highlighting another example on the HAR dataset, our method detects that for feature 50, client 16, a smaller weight is being applied. Investigating the underlying feature shows that, for this client, larger values are not being observed. While personalizing the local layers, \texttt{iFedAvg} therefore is reducing the value of this feature. This could be indicative of \textit{feature deactivation} or simply a compensation for the downstream effect of this feature in the \textit{shared} part of the model. The histogram of the feature and heatmap of $\ww_\inp$ can be seen in Figure \ref{fig:har_weight}.
    \begin{figure}
      \centering
      \includegraphics[width=0.99\linewidth]{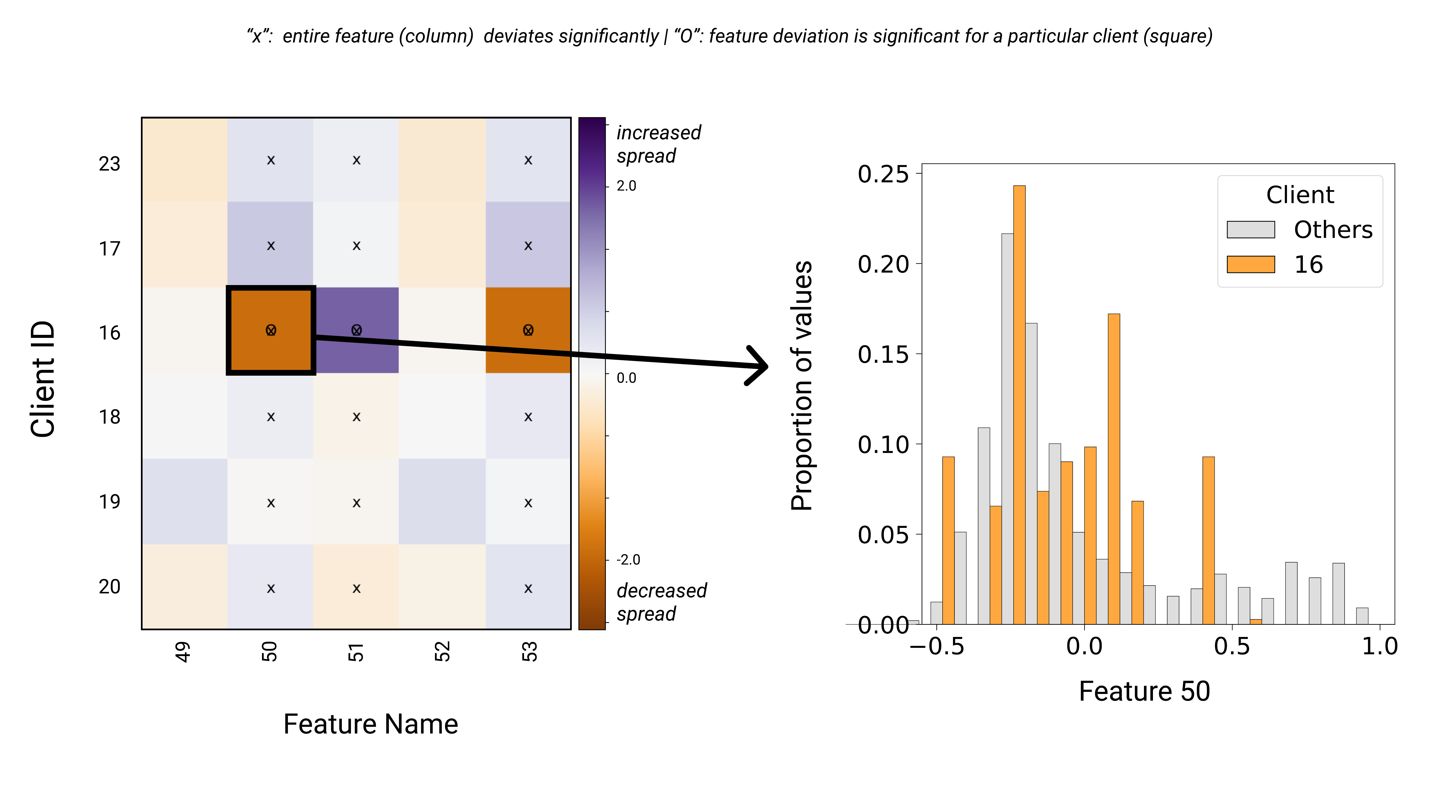}
      \caption{Heatmap of the weights ($\ww_\inp$) for the Human Activity Recognition dataset (left) and histograms of feature 50 for client 16 (right).}
      \label{fig:har_weight}
    \end{figure}
    
    \item \textbf{Vehicle Sensor Network - Target-specific Stretch:} A particularly interesting example of shifts detected by \texttt{iFedAvg} is one that is dependent on the target. For the VSN dataset, client 5 seems to have different values for feature 50 for the positive class, which is shown as significant in the heatmap. Client 19, albeit not significantly, has a slightly above average weight, and indeed for the negative target class has a differing underlying distribution. While the directionality of the weights in this instance do not directly indicate a shift for a particular target class, our method correctly identifies \textit{differences} in the underlying feature distribution. The heatmap and histograms of feature 50 for both clients and both target classes can be seen in Figure \ref{fig:vsn_weight}.
    \begin{figure}
      \centering
      \includegraphics[width=0.99\linewidth]{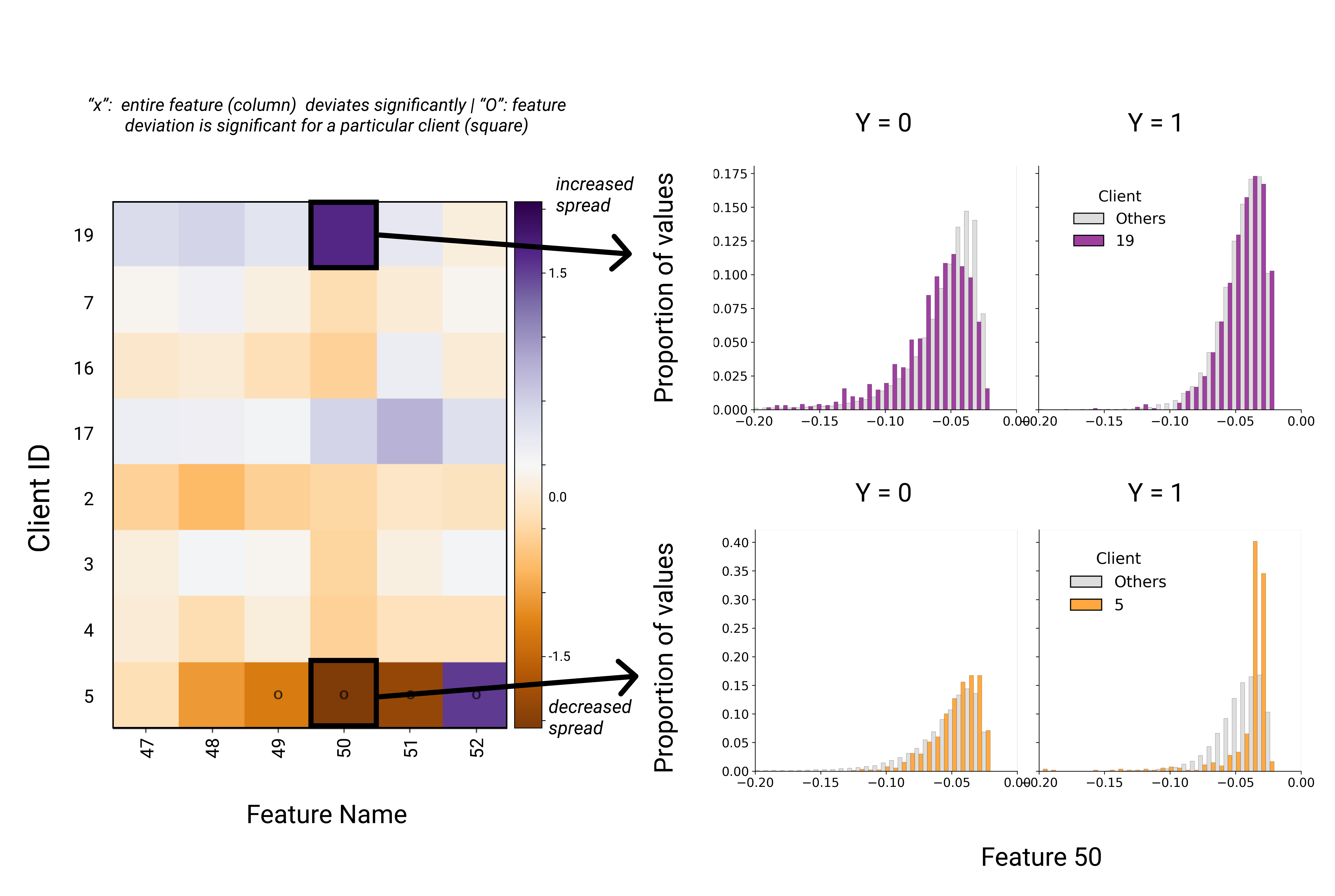}
      \caption{Heatmap of the weights ($\ww_\inp$) for the Vehicle Sensor Network dataset (left) and histograms of feature 50 for clients 19 and 5 split according to the target class (right).}
      \label{fig:vsn_weight}
    \end{figure}
    
    \item \textbf{Ebola Diagnosis - small underlying shifts:} Some of the discussed examples have been rather substantial and for features where many clients modify the bias or weight (marked by `x' in the columns). Detecting and correcting, small shifts can be critical for a client if they are the only ones performing such a personalized compensation. For the Kalihun ETC, it appears as if the referral times, the time taken until a patient actually visits the treatment center, are slightly larger and \texttt{iFedAvg} is compensating with a small but significant negative bias. For comparison, the histogram of an arbitrary ETC, Foya, is shown highlighting the difference in the underlying data. The heatmap and histograms are shown in Figure \ref{fig:diag_referral}
    \begin{figure}
      \centering
      \includegraphics[width=0.99\linewidth]{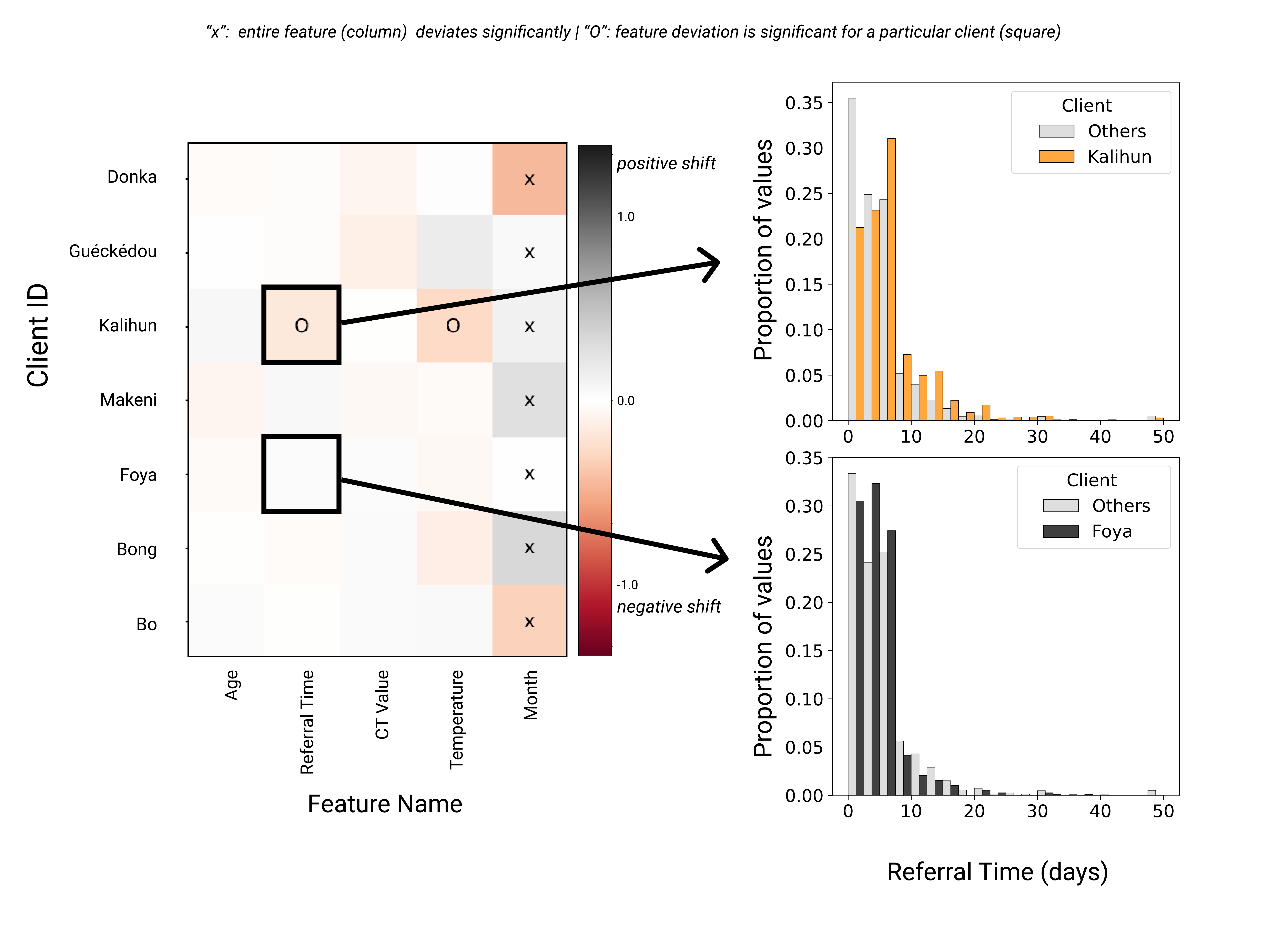}
      \caption{Heatmap of the biases ($\bb_\inp$) for the Ebola Diagnosis dataset (left) and histograms of the referral time feature for ETCs Kalihun and Foya (right).}
      \label{fig:diag_referral}
    \end{figure}

    \item \textbf{Ebola Diagnosis - data collection differences:} As \textit{bias} in the local data could have catastrophic consequences, we highlight another example. For ETC Foya, whether a patient has a malaria co-infection only appears to be recorded for EVD-positive cases. Here, the ETC only records a malaria test in confirmed cases. For comparison, another ETC, Kalihun, is shown, which does not record malaria infection at all. Without explicitly detecting this effect, the personalized model might overfit in practice, with detrimental consequences. The heatmap and both histograms are shown in Figure \ref{fig:diag_malaria}.
    \begin{figure}
      \centering
      \includegraphics[width=0.99\linewidth]{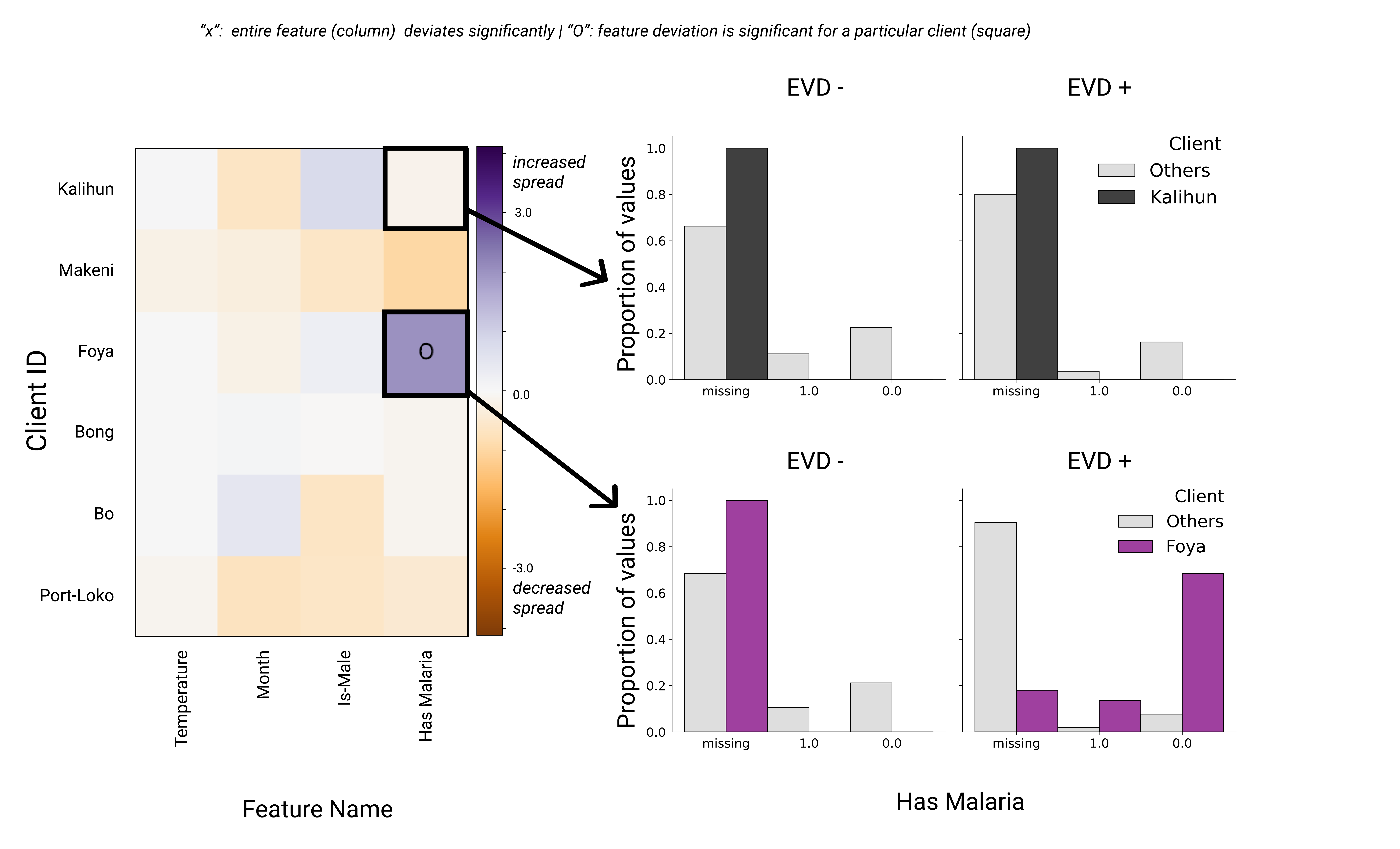}
      \caption{Heatmap of the weights ($\ww_\inp$) for the Ebola Diagnosis dataset (left) and histograms of the Malaria infection feature for ETCs Kalihun and Foya, split according to the target class (EVD- and EVD+) (right).}
      \label{fig:diag_malaria}
    \end{figure}
\end{itemize}

The previous examples show that \texttt{iFedAvg} is able to detect and compensate for various types of discrepancies in the underlying local datasets. Some, however, might not be desirable, and therefore being able to identify, at a client and feature level, problematic values could help reduce model bias. 

\newpage
\subsection{Hyperparameters and experimental setup}
\label{app:params_experiment}
In order to create the most comparable experiments, an identical network architecture was used for all experiments. This ensures that each method has the same number of parameters available in the base model. \texttt{APFL}, of course, creates multiple copies of this model. We show the entire MLP architecture in Table \ref{tab:architecture}, with $D$ the number of features and $K$ the output dimension. $K$ is chosen to be the number of classes in our experiments, after which log-softmax is applied in conjunction with negative-log-likelihood loss. The class weights, used to weight the loss function, are computed as the inverse of class prevalence, scaled to sum to $K$.

\begin{table}
  \caption{Architecture of the MLP model used}
  \label{tab:architecture}
  \centering
  \begin{tabular}{lccc}
    \toprule
    Type & Size (in, out) & Activation & Dropout \\
    \midrule
    $f_\inp$ & ($D$, $D$) & - & 0.2 \\
    Fully-connected & ($D$, 128) & TanH & 0.2 \\
    Fully-connected & (128, 64) & TanH & 0.2 \\
    Fully-connected & (64, $K$) & - & - \\
    $f_\out$ & ($K$, $K$) & - & - \\
    \bottomrule
  \end{tabular}
\end{table}

Each experiment was conducted on the following seeds, which dictate the local train and holdout set splitting, network initialization and batch shuffling. $2934384, 10231938, 8273, 2019231, 62739$. The learning rate of SGD was set to $0.002$ for all experiments and datasets, as this performed best across the board. The learning rate was decayed using a step function $50$ times, with a step of $0.9$ for the entirety of the $1000$ rounds. Client-side momentum with a value of $0.5$ was enabled for all methods except \texttt{APFL} as the authors do not discuss it. For \texttt{APFL} the best $\alpha$ was empirically found to be $0.5$. The F1 score was computed in a weighted fashion, ROC AUC with one-vs-one treatment for multi-class targets.

For both Ebola datasets, each client locally standardized the numerical features. For the VSN and HAR datasets, the original standardization of the benchmark datasets was retained. While there is no significant difference in the results, both modes are supported by \texttt{iFedAvg} and implemented in the opensourced code.

\newpage
\subsection{Target shift layer }
\label{app:target}

Intuitively, the layer $f_\out$ acts as a personalized compensation of any differences in how the target differs for each client. For instance, one would hope to detect varying class imbalance and a less clear class distinction in this layer. Interpreting this layer is slightly less intuitive, as each value corresponds to a logit, not a real feature. Nonetheless we present results in this section of \texttt{iFedAvg} with the training of local $f_\out$ enabled.

First, we analyze the performance with the target layer enabled. As can be seen in Table \ref{tab:targetperformance}, for most datasets there is a marginal performance gain. This difference is not significant enough to warrant this layer as necessary, but also highlights that it is not detrimental.
\begin{table}
  \caption{Performance difference with $f_\out$ enabled (F1 score)}
  \label{tab:targetperformance}
  \centering
  \begin{tabular}{lcc}
    \toprule
    Dataset & $\Delta$ Average & $\Delta$ Worst-performing client \\
    \midrule
    Ebola Prognosis & +0.002 & +0.008 \\
    Ebola Diagnosis & +0.005 & +0.008 \\
    Vehicle Sensor Network & +0.002 & - \\
    Human Activity Recognition & - & -0.013 \\
    \bottomrule
  \end{tabular}
\end{table}

We structure our analysis into the following three sections: 
\begin{itemize}
    \item \textbf{Overall results:} We show the heatmaps of $\bb_\out$ and $\ww_\out$ for EVD Diagnosis in Figure \ref{fig:targetlayer}. Each column now no longer represents input features, but the logit of each target class (0 being negative, 1 being positive diagnosis). In addition to the setting with both the bias and weight being trained locally, we also explore enabling the personalized training on each independently. It is visually apparent, that there are many smaller deviations, but few stand out except the weight of class 1 and ETC Foya. These shifts are not directly correlated with easily visible characteristics such as positivity rate. Therefore, the learned shifts are more complex to diagnose, and should be used as an indicator of potentially other issues. This is an interesting future research avenue.
    \begin{figure}
      \centering
      \includegraphics[width=0.99\linewidth]{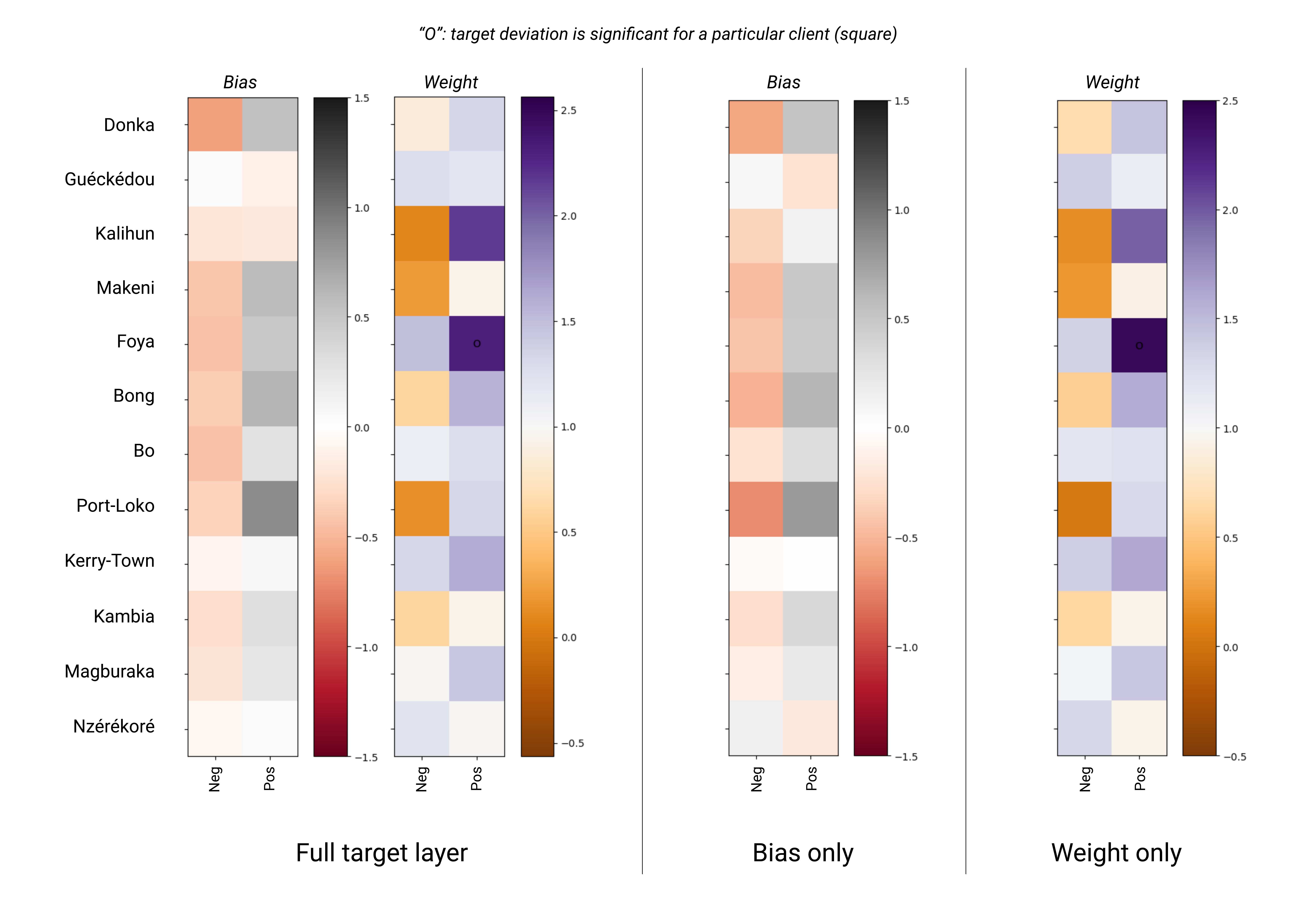}
      \caption{Heatmap of biases and weights ($\bb_\out$ and $\ww_\out$) of the target layer for the Ebola Diagnosis dataset. Each column represents the positive and negative class. Three configurations are shown: 1) both bias and weight trained locally (left), 2) only bias trained locally (middle) and only weight trained locally (right).}
      \label{fig:targetlayer}
    \end{figure}

    \item \textbf{Detecting mis-labeled targets:} While not as common as poorly standardized features, it is possible that due to a simple translation error, the label for a single client is flipped. With methods other than \texttt{iFedAvg}, this would result in terrible performance, and no interpretable output to help detect the issue. With an enabled target layer, $f_\out$, this becomes quite trivial.  Figure \ref{fig:targetflip} shows the heatmap for EVD Diagnosis, where for ETC Kalihun, the label was artificially swapped. An interesting alternative definition of $\ww_\out$ would be as a single scalar value, instead of a vector $\in \R^K$. This only allows each client a single multiplicative scaling of the last outputs. We show this setting in the right hand side of Figure \ref{fig:targetflip}. As can be observed in both constellations, a large negative weight is learned, indicating that for one class, the label is inverted. This is particularly apparent in the scalar setting. With this information, a client in the federation would have a strong indication that an interoperability issue does not originate from the features, but from the target itself. 
    \begin{figure}
      \centering
      \includegraphics[width=0.99\linewidth]{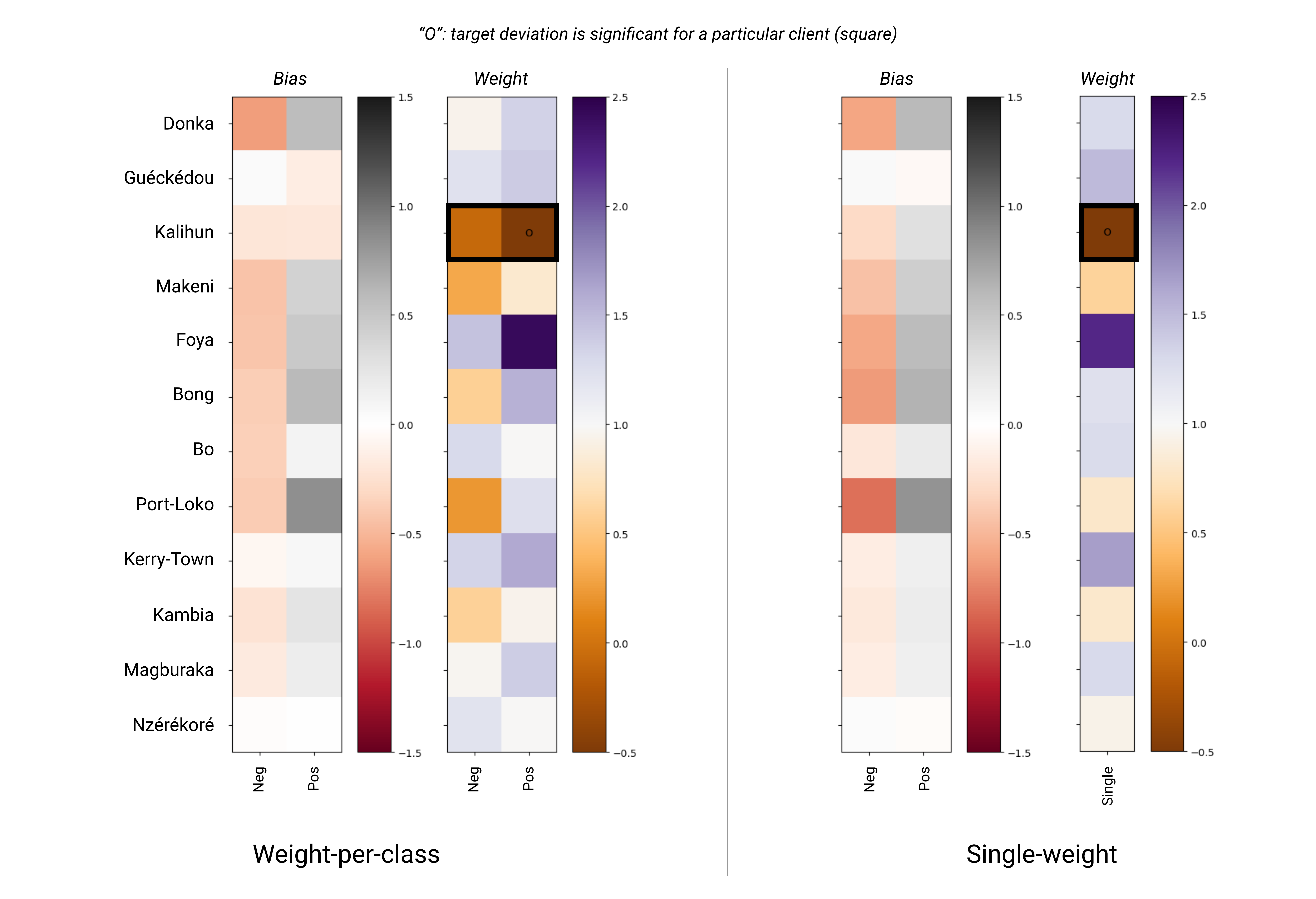}
      \caption{Heatmap of biases and weights ($\bb_\out$ and $\ww_\out$) of the target layer for the Ebola Diagnosis dataset with an artificially introduced target flip for ETC Kalihun (marked). Two configurations are shown: 1) $\ww_\out$ is a vector, with a value for each target class (left) and $\ww_\out$ is a scalar with a single value (right).}
      \label{fig:targetflip}
    \end{figure}
    
    \item \textbf{Consistent feature weights:} An undesirable effect of enabling the personalized target layer learning would be an impact on the feature-wise interpretable results. We highlight how this effect is marginal in Figure \ref{fig:targetimpact}. Therefore, the target layer can be enabled for most use-cases, especially when a target-swap could occur. 
    \begin{figure}
      \centering
      \includegraphics[width=0.99\linewidth]{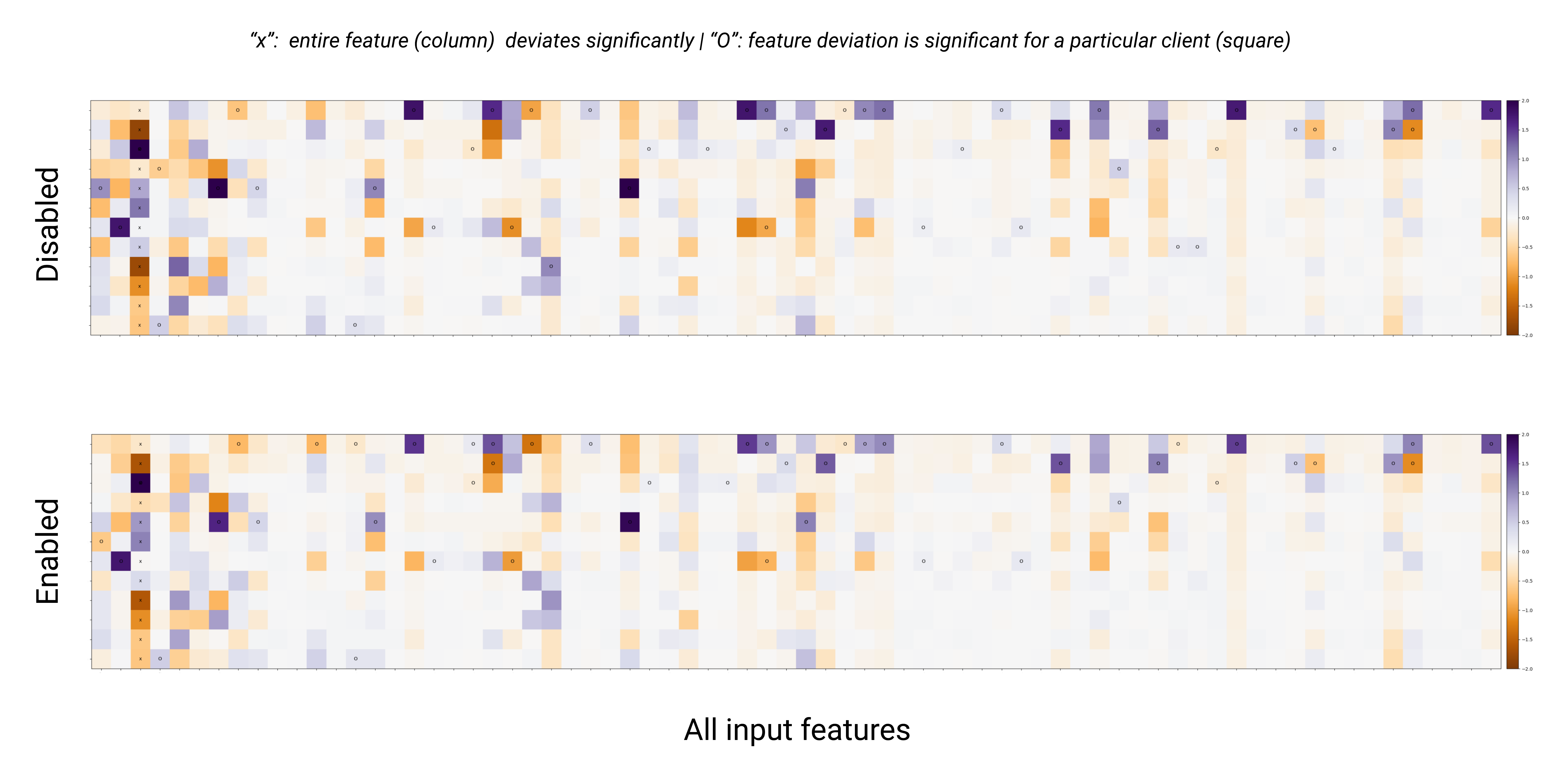}
      \caption{Heatmaps of the feature weights ($\ww_\inp$) for the Ebola Diagnosis dataset with local training of the target layer (bias and weight) disabled (top) and enabled (bottom)}
      \label{fig:targetimpact}
    \end{figure}
\end{itemize}

\newpage
\subsection{Ebola dataset statistics}
\label{app:ebola}

In Table \ref{tab:prognosis stats} and \ref{tab:diagnosis-stats} we show the number of samples at each treatment center. For prognosis, only patients where the outcome is known and a patient was confirmed EVD positive are considered. For diagnosis, only patients where an EVD test was performed and the minority class has at least 2\% of samples are included. This leads to the fact that not the same ETCs are represented for both tasks. The reason being that some ETCs did not monitor mortality, or others only treated EVD+ cases.

\begin{table}[H]
  \caption{Ebola Prognosis dataset summary statistics}
  \label{tab:prognosis stats}
  \centering
  \begin{tabular}{lcc}
    \toprule
    ETC & Number of samples & Mortality rate \\
    \midrule
    Guéckédou & 1366 & 66.98\% \\
    Monrovia &	1154 & 56.85\% \\
    Kalihun &	852 & 44.37\% \\
    Donka (EJPDEJ) & 748 & 49.87\% \\
    Foya &	450 & 66.00\% \\
    Bo &	440 & 38.63\% \\
    Donka (EFFVXT) & 418 & 37.80\% \\
    Kerry-Town &	263 & 42.59\% \\
    Port-Loko &	181 & 65.75\% \\
    Makeni &	176 & 56.82\% \\
    Bong &	168 & 50.00\% \\
    Freetown &	166 & 50.00\% \\
    \bottomrule
  \end{tabular}
\end{table}

\begin{table}
  \caption{Ebola Diagnosis dataset summary statistics}
  \label{tab:diagnosis-stats}
  \centering
  \begin{tabular}{lcc}
    \toprule
    ETC & Number of samples & Positivity rate \\
    \midrule
    Donka &	1975 & 37.87\% \\
    Guéckédou & 1517 & 90.05\% \\
    Kalihun &	1173 & 72.63\% \\
    Makeni &	848 & 20.75\% \\
    Foya &	564 & 79.79\% \\
    Bong &	529 & 31.76\% \\
    Bo &	519 & 84.78\% \\
    Port-Loko &	477 & 37.95\% \\
    Kerry-Town &	275 & 95.64\% \\
    Kambia &	217 & 21.66\% \\
    Magburaka &	155 & 29.03\% \\
    Nzérékoré & 137 & 57.66\% \\
    \bottomrule
  \end{tabular}
\end{table}

\end{document}